\documentclass[preprint]{elsarticle}
\usepackage{amsmath,amssymb,bm,graphicx,color}
\usepackage{lineno}
\usepackage[colorlinks, linkcolor=red, anchorcolor=blue, citecolor=green]{hyperref}
\usepackage[titletoc]{appendix}
\usepackage{subfigure}
\newcommand{\mbf}[1]{\mathbf{#1}}
\usepackage[table]{xcolor}
\definecolor{mygray}{gray}{.9}
\usepackage{adjustbox}
\usepackage{rotating}

\usepackage{algorithm}
\usepackage{algorithmic}

\journal{Mechanical Systems and Signal Processing}

\begin{document}
	
\begin{frontmatter}
		
\title{Deep Probabilistic Time Series Forecasting using Augmented Recurrent Input for Dynamic Systems}
		
\author[address1]{Haitao Liu}
\ead{htliu@dlut.edu.cn}

\author[address1]{Changjun Liu}
\ead{changjun.liu@mail.dlut.edu.cn}

\author[address1,address2]{Xiaomo Jiang}
\ead{xiaomojiang2019@dlut.edu.cn}

\author[address1]{Xudong Chen}
\ead{xdchen@mail.dlut.edu.cn}

\author[address1,address3]{Shuhua Yang}
\ead{yshemail@sina.com}

\author[address1]{Xiaofang Wang\corref{mycorrespondingauthor}}
\ead{dlwxf@dlut.edu.cn}

\cortext[mycorrespondingauthor]{Corresponding author}
\address[address1]{School of Energy and Power Engineering, Dalian University of Technology, China, 116024}
\address[address2]{Digital Twin Laboratory for Industrial Equipment, Dalian University of Technology, China, 116024.}
\address[address3]{Shenyang Blower Works Group Corporation, China, 110869.}

\begin{abstract}
The demand of probabilistic time series forecasting has been recently raised in various dynamic system scenarios, for example, system identification and prognostic and health management of machines. To this end, we combine the advances in both deep generative models and state space model (SSM) to come up with a novel, data-driven deep probabilistic sequence model. Specifically, we follow the popular encoder-decoder generative structure to build the recurrent neural networks (RNN) assisted variational sequence model on an augmented recurrent input space, which could induce rich stochastic sequence dependency. Besides, in order to alleviate the inconsistency issue of the posterior between training and predicting as well as improving the mining of dynamic patterns, we (i) propose using a lagged hybrid output as input for the posterior at next time step, which brings training and predicting into alignment; and (ii) further devise a generalized auto-regressive strategy that encodes all the historical dependencies for the posterior. Thereafter, we first investigate the methodological characteristics of the proposed deep probabilistic sequence model on toy cases, and then comprehensively demonstrate the superiority of our model against existing deep probabilistic SSM models through extensive numerical experiments on eight system identification benchmarks from various dynamic systems. Finally, we apply our sequence model to a real-world centrifugal compressor forecasting problem, and again verify its outstanding performance by quantifying the time series predictive distribution.
\end{abstract}

\begin{keyword}
State space model \sep Dynamic system \sep Recurrent neural networks \sep Variational inference \sep Variational autoencoder \sep Compressor
\end{keyword}

\end{frontmatter}


\section{Introduction}
Time series forecasting is a long standing problem in widespread decision-making scenarios, for example, dynamic system identification~\cite{aastrom1971system, verhaegen2007filtering}, prognostic and health management (PHM) of machines~\cite{zhao2019deep, yu2019remaining, lei2020applications}, and business demand forecasting~\cite{tadjer2021machine}. It thus has gained extensive attention from academic and industrial community over decades. 

In order to learn the underlying dynamic and temporal pattern from historical time series data and perform desirable forecasting in the future, classic time series methodologies, for example, exponential smoothing~\cite{hyndman2008forecasting}, and auto-regressive integrated moving average (ARIMA) model~\cite{box1968some, durbin2012time}, have been studied for a long history. They usually incorporate user's prior knowledge by decomposing the time series structure into trend, seasonality and so on. Consequently, these methods have high interpretability while suffering from poor capability of learning complex and long time dependency. Moreover, state space model (SSM)~\cite{hyndman2008forecasting, schon2011system, hanachi2016sequential}, which is similar to hidden Markov models~\cite{rabiner1986introduction} and could include the above classic methods as special cases~\cite{durbin2012time}, offers a more principled way to learn dynamic patterns from data.

For a forecasting model, an important feature is that it could not only provide the prediction mean but also quantify the associated uncertainty, which is essential for risk management, for instance, the estimation of remaining useful life~\cite{liao2018uncertainty}. To this end, an alternative way is interpreting the SSM paradigm from Bayesian view, for example, the well-known Kalman filter~\cite{kalman1960new} and the dynamic linear models~\cite{west2006bayesian}. Besides, it is popular to combine SSM with Gaussian process (GP)~\cite{williams2006gaussian}, which is theoretically equivalent to Kalman filter when using specific kernels. The Gaussian process state space model (GPSSM)~\cite{frigola2014variational, ialongo2019overcoming} places GP prior over both the transition and measurement functions of SSM, thus being capable of exporting probabilistic forecasting. The prominent weakness of GPSSM however is the poor scalability to tackle a large amount of time series data, due to the cubic time complexity~\cite{liu2020gaussian}. Though we could resort to the sparse and distributed approximations~\cite{snelson2006sparse, titsias2009variational, hensman2013gaussian, liu2018generalized} to improve the scalability of GPSSM, it on the other hand may limit the model capability due to the approximations.

Alternatively, the deep learning community has followed the SSM framework to extensively exploit the recurrent neural networks (RNNs), for example, gated recurrent unit (GRU)~\cite{cho2014learning}, long short term memory (LSTM)~\cite{hochreiter1997long} and their variants~\cite{sutskever2014sequence, gehring2017convolutional}, in order to perform time series \textit{point} prediction. Due to their high capability of extracting high-dimensional features for learning complicated patterns within time series data, the RNNs have reported remarkable and success stories in various domains~\cite{lipton2015critical, yu2021analysis}. In order to enable probabilistic forecasting using RNN, it is suggested to combine RNN with generative models. For example, we could combine the auto-regressive strategy with RNN and train the model through maximum likelihood~\cite{salinas2020deepar, li2021learning}. Alternatively, we could adopt RNN to learn the time-varying coefficients of linear/nonlinear stochastic SSM~\cite{rangapuram2018deep, yanchenko2020stanza}, thus enjoying both probabilistic forecasting and good interpretability. Another line of work builds the generative GP on the top of RNN as measurement function to produce stochastic outputs~\cite{al2017learning, she2020neural}. Finally, combining RNN with variational autoencoder (VAE)~\cite{kingma2013auto} has been exploited recently to build connections between SSM and VAE~\cite{bayer2014learning, krishnan2015deep, krishnan2017structured, fraccaro2018deep, li2021learning, gedon2020deep}. These models encode the sequence characteristics and dependency on the manifold embedded in a low-dimensional latent space, and are usually trained through variational inference (VI) by deriving the evidence lower bound (ELBO) as objective, thus resulting in the stochastic or variational RNN.

Along the line of deep SSM which gains benefits from two communities, we devise a novel, data-driven deep variational sequence model in an augmented recurrent input space, named VRNNaug, to perform probabilistic time series forecasting. The main contributions of this paper are three-fold.
\begin{itemize}
	\item We propose a VAE-type sequence model on an augmented input space, the inputs of which consist of the original input signal and the additional latent input which encodes all the sequence characteristics. The augmented input space could induce rich stochastic sequence dependency.
	\item We alleviate the inconsistency issue of the posterior between training and predicting as well as improving the modeling of dynamic patterns in the VI framework by (i) feeding the lagged hybrid output as input for the posterior at next time step, which brings training and predicting into alignment; and (ii) presenting a generalized auto-regressive strategy which encodes all the historical dependencies for the posterior. This is found to greatly improve the quality of future forecasting.
	\item We perform extensive numerical experiments to verify the superiority of our model on eight system identification benchmarks from various dynamic systems, and thereafter apply the proposed sequence model on a real-world centrifugal compressor case to illustrate the desirable time series forecasting.
\end{itemize}

The remainder of this paper is organized as follows. Section~\ref{sec_preliminary} first defines the time series forecasting problem, and then section~\ref{sec_method} presents the proposed deep variational sequence model based on augmented recurrent input space. Thereafter, section~\ref{sec_exp} conducts a comprehensive comparison study on eight system identification benchmarks as well as a centrifugal compressor case to verify the superiority of proposed sequence model. Finally, concluding remarks of this paper are provided in section~\ref{sec_remarks}. Note that the acronyms and notations used in this paper are summarized in Appendices~\ref{app_acronyms} and~\ref{app_notation}.

\section{Preliminaries} \label{sec_preliminary}
It is assumed that the sequence data $\mathcal{D} = \{\mbf{u}_{1:T} \in \mathbb{R}^{T \times d_{\mbf{u}}}, \mbf{y}_{1:T} \in \mathbb{R}^{T \times d_{\mbf{y}}} \}$ at $T$ discrete and evenly distributed time points are sampled from an unknown dynamic system. We attempt to model it in order to learn the underlying dynamic and temporal patterns. Thereafter, we use the fitted sequence model $\mathcal{M}$ to forecast the predictive distribution $p(\mbf{y}_{T+1:T+F}|\mathcal{D}, \mbf{u}_{T+1:T+F})$ of future window $\mbf{y}_{T+1:T+F}$ given the history data $\mathcal{D}$, the forecast horizon $F$, and the further input signals (also known as covariates or exogenous variables) $\mbf{u}_{T+1:T+F}$.

Specifically, we assume that the target dynamic system could be described by the state space model as
\begin{align}
	\mbf{h}_t =& f (\mbf{h}_{t-1}, \mbf{u}_t), \label{eq_ssm_1} \\
	\mbf{y}_t =& g (\mbf{h}_t), \label{eq_ssm_2}
\end{align}
where the subscript $t$ $(1 \le t \le T)$, which is a non-negative integer, indexes time; the variable $\mbf{h}_t \in \mathbb{R}^{d_{\mbf{h}}}$ is the system hidden state to be inferred at time $t$ and $\mbf{h}_0$ represents the initial hidden state; the input signal (action) $\mbf{u}_t$ is known at all time points; $\mbf{y}_t \in \mathbb{R}^{d_{\mbf{y}}}$ is the measurement (observation) at time $t$; the transition function $f: \mathbb{R}^{d_{\mbf{h}}} \mapsto \mathbb{R}^{d_{\mbf{h}}}$ evolves the hidden state $\mbf{h}_{t-1}$ to the next state $\mbf{h}_{t}$;\footnote{When the transition $f$ is a linear time-invariant mapping, we arrive at the linear SSM.} and the measurement function (also known as emission function) $g: \mathbb{R}^{d_{\mbf{h}}} \mapsto \mathbb{R}^{d_{\mbf{y}}}$ maps the hidden state to the associated output.

It is observed that when we are describing the functions $f$ and $g$ using deep mappings, e.g., neural networks, it results in the so-called deep SSM which takes the advances from two fields: the high flexibility from neural networks and the high expressivity via hidden states. When future combining SSM with generative neural networks, for example, VAE, we arrive at the subclass of deep SSM, which is capable of performing probabilistic time series forecasting and will be investigated below in our work. Note that the possible perturbations, for example, modeling error and measurement error, have been absorbed in~\eqref{eq_ssm_1} and~\eqref{eq_ssm_2}.

\section{Variational sequence model using augmented stochastic inputs} \label{sec_method}
In order to incorporate stochasticity into the sequence model, we propose to use additional random variable to augment the input space as $\mbf{u}_t \in \mathbb{R}^{d_{\mbf{u}}} \mapsto [\mbf{u}_t, \mbf{z}_t] \in \mathbb{R}^{d_{\mbf{u}} + d_{\mbf{z}}}$. Consequently, rich stochastic sequence representation could be induced in the augmented input space for improving both the model capability and the quantification of uncertainty. The proposed model as well as the inference and implementation details will be elaborated as below.

\subsection{Model definition}
Under the framework of latent variable model, the joint prior with additional latent variables $\mbf{z}_{1:T}$ is assumed to factorize over time as
\begin{align} \label{eq_p_z}
\begin{aligned}
	p(\mbf{y}_{1:T}, \mbf{z}_{1:T} | \mbf{u}_{1:T}) =& \prod_{t=1}^T p(\mbf{y}_t|\mbf{z}_t, \mbf{u}_t) p(\mbf{z}_t|\mbf{z}_{0:t-1}, \mbf{u}_{1:t}).
\end{aligned}
\end{align}
In~\eqref{eq_p_z}, the latent sequence $\mbf{z}$ starts from the initial input state $\mbf{z}_0$ which could take for example the zero initialization for cold starting. It is observed that the Gaussian state transition $p(\mbf{z}_t|\mbf{z}_{0:t-1}, \mbf{u}_{1:t})$ $(1 \le t \le T)$ (i) evolves over time for encoding the sequence dependency; and (ii) it is particularly conditioned on all the historical input signals $\mbf{u}_{1:t}$ until time $t$ and all the historical latent variables $\mbf{z}_{0:t-1}$ until time $t-1$, not only the information limited at adjacent time points $t-1$ and $t$. The consideration of all the historical information in $p(\mbf{z}_t|.)$ is expected to help learn long sequence patterns. Besides, the Gaussian likelihood $p(\mbf{y}_t|\mbf{z}_t, \mbf{u}_t)$ in~\eqref{eq_p_z} plays as the role of decoder for mapping the encoded sequence inputs to the related output at time $t$.\footnote{Note that the likelihood is not limited to Gaussian, it can adopt other task-dependent distributions.} This joint prior can be estimated by Monte Carlo sampling conducted recursively over time.

For model inference, we employ the type-II maximum likelihood to tune model parameters by maximizing the marginal likelihood
\begin{align}
p(\mbf{y}_{1:T}| \mbf{u}_{1:T}) = \int p(\mbf{y}_{1:T}, \mbf{z}_{1:T} | \mbf{u}_{1:T}) d\mbf{z}_{1:T}.
\end{align}
When performing prediction, we require the posterior $p(\mbf{z}_{1:T} | \mbf{u}_{1:T}, \mbf{y}_{1:T})$, the distribution of which however is unknown. We hence resort to variational inference by introducing Gaussian variational posterior as an approximation. Specifically, similar to the prior $p(\mbf{z}_{1:T} | \mbf{u}_{1:T}) = \prod_{t=1}^T p(\mbf{z}_t|\mbf{z}_{0:t-1}, \mbf{u}_{1:t}) $ in~\eqref{eq_p_z}, the variational posterior could be factorized over time as
\begin{align} \label{eq_q_z}
\begin{aligned}
	q(\mbf{z}_{1:T} | \mbf{u}_{1:T}, \mbf{y}_{1:T}) =& \prod_{t=1}^T q(\mbf{z}_t|\mbf{z}_{0:t-1}, \mbf{u}_{1:t}, \mbf{y}_{1:t}),
\end{aligned}
\end{align}
which describes the generation of augmented latent states given observations and input signals up to current time, and each state transition $q(\mbf{z}_t|\mbf{z}_{0:t-1}, \mbf{u}_{1:t}, \mbf{y}_{1:t})$ takes the Gaussian form. 

Again, it is worth noting that the additional encoding of all the previous variables in~\eqref{eq_p_z} and~\eqref{eq_q_z} enables the learning of long term dependency. The encoding method using generalized auto-regressive strategy will be detailed in section~\ref{sec_amortized_model}, and its superiority over the usage of only adjacent variables will be verified in the ablation study in section~\ref{sec_ablation}.

\subsection{Derivation of ELBO} 
To simultaneously infer the interested marginal likelihood $p(\mbf{y}_{1:T}| \mbf{u}_{1:T})$ and the variational posterior $q(\mbf{z}_{1:T} | \mbf{u}_{1:T}, \mbf{y}_{1:T})$, we resort to variational inference. To this end, we minimize the Kullback-Leibler (KL) divergence between the variational posterior $q(\mbf{z}_{1:T} | \mbf{u}_{1:T}, \mbf{y}_{1:T})$ and the exact posterior $p(\mbf{z}_{1:T} | \mbf{u}_{1:T}, \mbf{y}_{1:T})$, which is formulated as
\begin{align} \label{eq_KL_z}
\begin{aligned}
& \mathrm{KL}[q(\mbf{z}_{1:T} | \mbf{u}_{1:T}, \mbf{y}_{1:T}) || p(\mbf{z}_{1:T} | \mbf{u}_{1:T}, \mbf{y}_{1:T})] \\
=& \mathbb{E}_{q(\mbf{z}_{1:T} | \mbf{u}_{1:T}, \mbf{y}_{1:T})} \left[ \log \frac{q(\mbf{z}_{1:T} | \mbf{u}_{1:T}, \mbf{y}_{1:T})}{p(\mbf{z}_{1:T} | \mbf{u}_{1:T}, \mbf{y}_{1:T})} \right] \\
=& \log p(\mbf{y}_{1:T}| \mbf{u}_{1:T}) + \mathbb{E}_{q(\mbf{z}_{1:T} | \mbf{u}_{1:T}, \mbf{y}_{1:T})} \left[ \log \frac{q(\mbf{z}_{1:T} | \mbf{u}_{1:T}, \mbf{y}_{1:T})}{p(\mbf{z}_{1:T}, \mbf{y}_{1:T}| \mbf{u}_{1:T})} \right].
\end{aligned}
\end{align}
Since the KL term~\eqref{eq_KL_z} is always non-negative, we derive the following ELBO for the log marginal likelihood $\log p(\mbf{y}_{1:T}|\mbf{u}_{1:T})$ as
\begin{align} 
\begin{aligned}
\log p(\mbf{y}_{1:T}|\mbf{u}_{1:T}) \ge \mathcal{L} =& \mathbb{E}_{q(\mbf{z}_{1:T}| \mbf{u}_{1:T}, \mbf{y}_{1:T})} \left[ \log \frac{p(\mbf{y}_{1:T}, \mbf{z}_{1:T} | \mbf{u}_{1:T})}{q(\mbf{z}_{1:T} | \mbf{u}_{1:T}, \mbf{y}_{1:T})} \right].
\end{aligned}
\end{align}
By inserting the joint prior~\eqref{eq_p_z} and the variational posterior~\eqref{eq_q_z} into $\mathcal{L}$, we arrive at the factorization over time as
\begin{align}
\begin{aligned} \label{eq_elbo}
\mathcal{L}=& \sum_{t=1}^T \mathbb{E}_{q(\mbf{z}_{t}|\mbf{z}_{0:t-1}, \mbf{u}_{1:t}, \mbf{y}_{1:t})} [\log p(\mbf{y}_t|\mbf{z}_t, \mbf{u}_t)] \\
&- \mathrm{KL}[q(\mbf{z}_t|\mbf{z}_{0:t-1}, \mbf{u}_{1:t}, \mbf{y}_{1:t}) || p(\mbf{z}_t|\mbf{z}_{0:t-1}, \mbf{u}_{1:t})],
\end{aligned}
\end{align}
where the latent prior usually takes the unit Gaussian $p(\mbf{z}_t|\mbf{z}_{0:t-1}, \mbf{u}_{1:t}) = \mathcal{N}(\mbf{z}_t|\mbf{0}, \mbf{I})$ for simplicity; and for the likelihoods and variational posteriors, all of them take the Gaussian form amortized through neural networks, which will be elaborated in section~\ref{sec_amortized_model}. Besides, given the Gaussian posterior $q(\mbf{z}_t|\mbf{z}_{0:t-1}, \mbf{u}_{1:t}, \mbf{y}_{1:t}) = \mathcal{N}(\mbf{z}_t|\bm{\mu}_{\mbf{z}_t}, \bm{\Sigma}_{\mbf{z}_t})$ with the diagonalized variance $\bm{\Sigma}_{\mbf{z}_t} = \mathrm{diag}[\bm{\nu}_{\mbf{z}_t}]$, the KL terms in the right-hand side of~\eqref{eq_elbo} have analytical expressions as
\begin{align}
\begin{aligned}
&\mathrm{KL}[q(\mbf{z}_t|\mbf{z}_{0:t-1}, \mbf{u}_{1:t}, \mbf{y}_{1:t}) || p(\mbf{z}_t|\mbf{z}_{0:t-1}, \mbf{u}_{1:t})] \\
=& \frac{1}{2} \left[\mathrm{Tr}(\bm{\Sigma}_{\mbf{z}_t}) + \bm{\mu}_{\mbf{z}_t}^{\mathsf{T}}\bm{\mu}_{\mbf{z}_t} - d_{\mbf{z}} - \log|\bm{\Sigma}_{\mbf{z}_t}| \right].
\end{aligned}
\end{align}
Finally, it is found that minimizing the KL divergence in~\eqref{eq_KL_z} is equivalent to maximizing the ELBO in~\eqref{eq_elbo}, which (i) pushes the bound $\mathcal{L}$ towards the marginal likelihood and (ii) infers the interested variational posterior $q(\mbf{z}_t|.)$.

The estimation of ELBO $\mathcal{L}$, especially the first likelihood expectation in the right-hand side of~\eqref{eq_elbo}, could be conducted through the re-parameterization trick~\cite{kingma2013auto} due to the latent Gaussian random variables, therefore resulting in a differentiable objective for optimization. For instance, instead of directly sampling from the variational Gaussian posterior $q(\mbf{z}_t|.)$, we sample from another random variable $\bm{\epsilon} \sim \mathcal{N}(\mbf{0}, \mbf{I})$ without trainable parameters, and then use it to obtain $\hat{\mbf{z}}_t = \bm{\mu}_{\mbf{z}_t} + \sqrt{\bm{\nu}_{\mbf{z}_t}} \odot \bm{\epsilon}$ where the symbol $\odot$ represents element-wise product. 

Besides, directly evaluating the recurrent ELBO $\mathcal{L}$ unrolled over a long time series requires high memory consumption due to the many historical states. More efficiently, we employ the \textit{shingling} technique proposed in~\cite{leskovec2014mining} to convert the long time series into several short time series chunks. Specifically, given a long time series $\mathcal{D}$ with length of $T$, we start from an arbitrary time index $t$ and obtain a short time series with sequence length $W$ ($W < T$). Consequently, we could obtain $J = T - W + 1$ short time series chunks $\mathcal{D}_c = \{\mathcal{D}_c^j\}_{j=1}^{J}$, each of which has the sequence length of $W$. This shingling data preprocessing allows modeling many independent time series with varying length simultaneously. Besides, the above time series split could naturally support an efficient and unbiased estimation of $\mathcal{L}$ on a subset $\mathcal{B} \subseteq \mathcal{D}_c$ of the short time series chunks with $|\mathcal{B}| \ll J$, thus allowing using mini-batch optimizer, e.g., Adam~\cite{kingma2014adam}. That is, an unbiased estimation of $\mathcal{L}$ on the subset $\mathcal{B}$ can be expressed as
\begin{align} \label{eq_elbo_unbiased}
\tilde{\mathcal{L}} = \frac{J}{|\mathcal{B}|} \sum_{\mathcal{D}_c^j \in \mathcal{B}} \mathcal{L}^j,
\end{align}
where $\mathcal{L}^j$ indicates the ELBO evaluated on the short time series $\mathcal{D}_c^j$ via~\eqref{eq_elbo}.

\subsection{Amortized VRNNaug} \label{sec_amortized_model}
\begin{figure}[t!]
	\centering
	\includegraphics[width=0.7\textwidth]{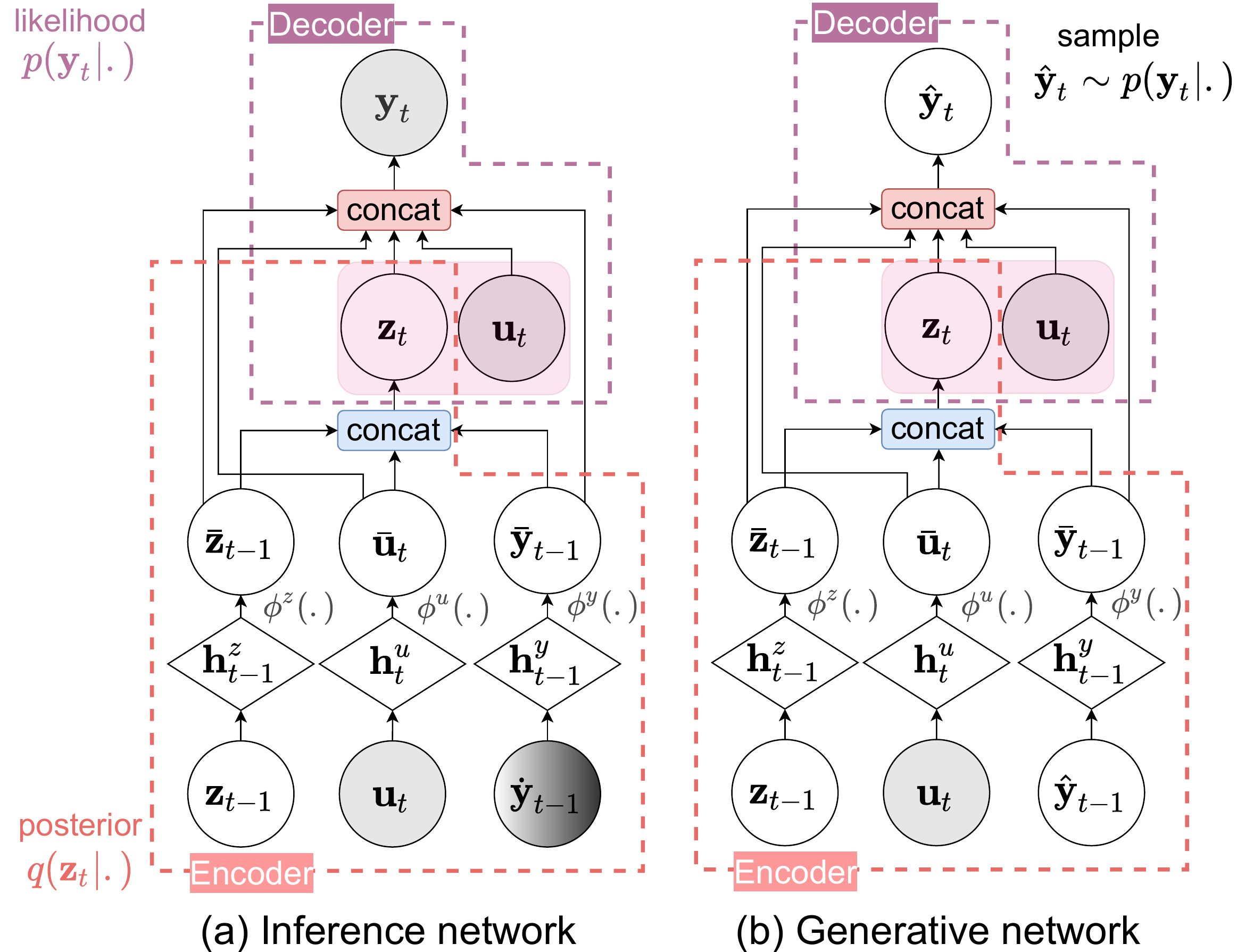}
	\caption{The inference and generative networks of the proposed VRNNaug model for probabilistic time series forecasting. Note that the gray nodes represent input signals and observations. Particularly, the transient gray node $\dot{\mbf{y}}_{t-1}$ indicates the mixture of observation $\mbf{y}_{t-1}$ and sample $\hat{\mbf{y}}_{t-1}$. Finally, it is found that in comparison to the inference network, the major difference in the generative network is that it is using the sample $\hat{\mbf{y}}_{t-1}$ since we are inaccessible to the true observation when predicting.}
	\label{fig_vrnnaug}
\end{figure}

After deriving the ELBO~\eqref{eq_elbo} for the proposed VRNNaug, the next is to elaborate the practical implementations for constructing the variational posterior $q(\mbf{z}_t|.)$ and the likelihood $p(\mbf{y}_t|.)$. The detailed implementations of the proposed VRNNaug sequence model are depicted in Fig.~\ref{fig_vrnnaug}. It is observed that the whole model follows the encoder-decoder structure, wherein the encoder embeds the temporal patterns into the stochastic, time-aware manifold represented by $\mbf{z}_t$, while the decoder generates outputs from the time-aware manifold. Furthermore, we employ amortized variational inference, i.e., amortizing the parameters over neural networks, to build and train our proposed VRNNaug model efficiently. Particularly, (i) a hybrid strategy is utilized to address the inconsistency of $q(\mbf{z}_t|.)$ between training and predicting, and (ii) a generalized auto-regressive strategy is proposed in building $q(\mbf{z}_t|.)$ in order to summarize the long historical patterns, which will be elaborated in what follows.

\textbf{Inconsistency issue.} Before proceeding to the amortized modeling, we first discuss the inconsistency issue occurred in the proposed model. It is found that the stochastic sequence model defined in~\eqref{eq_p_z} and~\eqref{eq_q_z} however may raise \textit{inconsistency} between training and predicting when we resort to the variational posterior $q(\mbf{z}_t|\mbf{z}_{0:t-1}, \mbf{u}_{1:t}, \mbf{y}_{1:t})$. That is, different from the training phase, when we are going to predict at the future time $T+1$, we have to use the prior $p(\mbf{z}_{T+1}|\mbf{z}_{0:T}, \mbf{u}_{1:T+1})$ instead of the variational posterior $q(\mbf{z}_{T+1}|\mbf{z}_{0:T}, \mbf{u}_{1:T+1}, \mbf{y}_{1:T+1})$, since there is no access to the further observation $\mbf{y}_{T+1}$. The gap raised by this inconsistency might deteriorate the quality of forecasting. 

Inspired by~\cite{salinas2020deepar}, in order to alleviate the inconsistency issue, we could use the \textit{lagged} output $\mbf{y}_{t-1}$ instead of $\mbf{y}_t$ in the conditions of the variational posterior $q(\mbf{z}_t|.)$ with $t \le T$ during the training phase. Besides, we sample $\hat{\mbf{y}}_{t-1} \sim p(\mbf{y}_{t-1}|.)$ from the previous probabilistic output when predicting at time $t$ with $T+1 \le t \le T+F$. This auto-regressive strategy extracts the known previous output or sample and feeds it as input for $q(\mbf{z}_t|.)$ at next time point, which is available for both training and predicting. However, this strategy introduces another inconsistency: we use the ground truth $\mbf{y}_{t-1}$ during training, but the sample $\hat{\mbf{y}}_{t-1}$ during predicting. This discrepancy has been found to raise error accumulation, thus deteriorating the RNN performance on natural language processing (NLP) tasks~\cite{bengio2015scheduled}. 

Therefore, inspired by~\cite{zhang2019bridging}, we decide to use a simple trick to alleviate the discrepancy by taking the following lagged \textit{hybrid}
\begin{align} \label{eq_hyb_y}
	\dot{\mbf{y}}_{t-1} = \frac{1}{2}(\mbf{y}_{t-1} + \hat{\mbf{y}}_{t-1}), \quad 1 \le t \le T,
\end{align}
and feeding it as input at next time point for the posterior $q(\mbf{z}_t|.)$ during training, see Fig.~\ref{fig_vrnnaug}(a). The ground truth $\mbf{y}_{t-1}$ in~\eqref{eq_hyb_y} plays as the teaching force to accelerate model convergence, while the usage of predictive sample $\hat{\mbf{y}}_{t-1}$ plays as the regularizer to reduce the gap between training and predicting. Besides, akin to $\mbf{z}_0$, the initials $\mbf{y}_0$ and $\hat{\mbf{y}}_0$ in~\eqref{eq_hyb_y} could simply take zero initialization for cold starting, thus resulting in $\dot{\mbf{y}}_0 = \mbf{0}$. Note that when the learned output distributions are close to the true distributions of observations in the later stage of training, we have $\dot{\mbf{y}}_{t-1} \approx \mbf{y}_{t-1}$.

\textbf{NN-assisted modeling.} To perform amortized variational inference, we attempt to parameterize the Gaussian variational posterior $q(\mbf{z}_t|.)$ and the Gaussian likelihood $p(\mbf{y}_t|.)$ in ELBO~\eqref{eq_elbo} through neural networks. Firstly, the mean $\bm{\mu}_{\mbf{z}_t}$ and variance $\bm{\nu}_{\mbf{z}_t}$ of $q(\mbf{z}_t|.)$ depend on the historical data $\mbf{z}_{0:t-1}$, $\mbf{u}_{1:t}$ and $\dot{\mbf{y}}_{0:t-1}$, and are parameterized through a multi-layer perceptron (MLP) as
\begin{align} \label{eq_q_zt}
\begin{aligned}
\left[\bm{\mu}_{\mbf{z}_t},\log\bm{\nu}_{\mbf{z}_t}\right] =& f^z(\mbf{z}_{0:t-1}, \mbf{u}_{1:t}, \dot{\mbf{y}}_{0:t-1}) = \mathrm{MLP}^z(\bar{\mbf{z}}_{t-1}, \bar{\mbf{u}}_{t}, \bar{\mbf{y}}_{t-1})
\end{aligned}
\end{align}
In~\eqref{eq_q_zt}, the recurrent transformations $\bar{\mbf{z}}_{t-1} \in \mathbb{R}^{d_{\mbf{z}}}$, $\bar{\mbf{u}}_{t} \in \mathbb{R}^{d_{\mbf{u}}}$ and $\bar{\mbf{y}}_{t-1} \in \mathbb{R}^{d_{\mbf{y}}}$ respectively employ the \textit{generalized} auto-regressive strategy to evolve over all the historical states $\mbf{z}_{0:t-1}$, $\mbf{u}_{1:t}$ and $\dot{\mbf{y}}_{0:t-1}$ as
\begin{align}
\bar{\mbf{z}}_{t-1} =& \phi^z(\mbf{z}_{0:t-1}), \label{eq_bar_z} \\ 
\bar{\mbf{u}}_{t} =& \phi^u(\mbf{u}_{1:t}), \label{eq_bar_u} \\ 
\bar{\mbf{y}}_{t-1} =& \phi^y(\dot{\mbf{y}}_{0:t-1}), \label{eq_bar_y} 
\end{align}
in order to extract long temporal patterns. Specifically, the recurrent mappings $\phi^z(.)$, $\phi^u(.)$ and $\phi^y(.)$ shown in Fig.~\ref{fig_vrnnaug} are achieved by building upon the outputs of for example RNN models as
\begin{align}
	\mbf{h}^z_{t-1} &= \mathrm{RNN}^{z}(\mbf{h}^z_{t-2}, \mbf{z}_{t-1}), \quad \bar{\mbf{z}}_{t-1} = \mathrm{MLP}^{\bar{z}}(\mbf{h}^z_{t-1}), \label{eq_ar_z} \\
	\mbf{h}^u_{t} &= \mathrm{RNN}^u(\mbf{h}^u_{t-1}, \mbf{u}_{t}), \quad \bar{\mbf{u}}_{t} = \mathrm{MLP}^{\bar{u}}(\mbf{h}^u_{t}), \label{eq_ar_u} \\
	\mbf{h}^y_{t-1} &= \mathrm{RNN}^y(\mbf{h}^y_{t-2}, \dot{\mbf{y}}_{t-1}), \quad \bar{\mbf{y}}_{t-1} = \mathrm{MLP}^{\bar{y}}(\mbf{h}^y_{t-1}). \label{eq_ar_y}
\end{align}
The above RNN cells, for instance, GRU~\cite{cho2014learning} and LSTM~\cite{hochreiter1997long}, naturally encode the sequence characteristics. It is notable that the recurrent mappings $\phi^z(.)$, $\phi^u(.)$ and $\phi^y(.)$ are not limited to RNN models. The recently developed sequence extractors, for example, temporal convolutional networks (TCN)~\cite{bai2018empirical} and transformer~\cite{vaswani2017attention}, are also available here.

Besides, it is worth noting that the recurrent transformation $\bar{\mbf{y}}_{t-1}$ in~\eqref{eq_bar_y} and the hidden state $\mbf{h}^y_{t-1}$ in~\eqref{eq_ar_y} solely hold for training, see the inference network in Fig.~\ref{fig_vrnnaug}(a). When predicting, we have $\bar{\mbf{y}}_{t-1} = \phi^y(\hat{\mbf{y}}_{0:t-1})$ that accepts the samples from previous output distributions, with the hidden state now formulated as
\begin{align} \label{eq_ar_y_hat}
 \mbf{h}^y_{t-1} = \mathrm{RNN}^y(\mbf{h}^y_{t-2}, \hat{\mbf{y}}_{t-1}),
\end{align}
see the generative network in Fig.~\ref{fig_vrnnaug}(b).

Secondly, as for the Gaussian likelihood (decoder) in the ELBO~\eqref{eq_elbo}, it writes as
\begin{align} \label{eq_p_yt}
	p(\mbf{y}_t|\mbf{z}_t, \mbf{u}_t) = \mathcal{N}(\mbf{y}_t|\bm{\mu}_{\mbf{y}_t}, \bm{\Sigma}_{\mbf{y}_t}),
\end{align}
where the diagonalized variance $\bm{\Sigma}_{\mbf{y}_t} = \mathrm{diag}[\bm{\nu}_{\mbf{y}_t}]$. Since $\mbf{z}_t$ is mapped from the recurrent inputs $\bar{\mbf{z}}_{t-1}$, $\bar{\mbf{u}}_{t}$ and $\bar{\mbf{y}}_{t-1}$ via $\phi^z(.)$, $\phi^u(.)$ and $\phi^y(.)$, we follow the architecture of Dense-Net~\cite{huang2017densely} by extracting and concatenating all the previous inputs in Fig.~\ref{fig_vrnnaug} in order to parameterize the Gaussian parameters through MLP as
\begin{align} \label{eq_mu_nu_yt}
	[\bm{\mu}_{\mbf{y}_t},\log\bm{\nu}_{\mbf{y}_t}] = \mathrm{MLP}^y(\mbf{z}_t, \mbf{u}_t, \bar{\mbf{z}}_{t-1}, \bar{\mbf{u}}_{t}, \bar{\mbf{y}}_{t-1}).
\end{align}
This sort of \textit{dense connection} between encoder and decoder eases model training because of the block-cross feature sharing. Finally, we again highlight that the likelihood is not limited to Gaussian, it can adopt other task-dependent distributions. For instance, we could choose the negative-binomial likelihood for positive count data, and the Bernoulli likelihood for binary data~\cite{salinas2020deepar}. 

\subsection{Prediction} 
As shown in Fig.~\ref{fig_vrnnaug}(b), given the trained deep sequence model $\mathcal{M}$, in order to perform recursive prediction over the future time $t \in [T+1, T+F]$ with forecast horizon $F$, we first recursively sample $\hat{\mbf{z}}_t$ from the state transition $q(\mbf{z}_t|.)$. Note that the recurrent transformation $\bar{\mbf{y}}_{t-1}$ in~\eqref{eq_q_zt} now is unrolled over the previous output samples $\hat{\mbf{y}}_{0:t-1}$ via~\eqref{eq_ar_y_hat} since the true output observations are inaccessible when predicting. Thereafter, we pass the latent sample through the decoder $p(\mbf{y}_t|.)$ to output the prediction sample $\hat{\mbf{y}}_t$ at current time point. It is notable that although the model employs Gaussian variables, the final predictive distribution is not Gaussian. Hence, we could use Monte Carlo sampling to output $K$ samples for empirically quantifying the implicit predictive distribution. 

Finally, pseudo implementations of the proposed VRNNaug model for training and predicting are provided in Algorithms~\ref{alg_vrnnaug_train} and~\ref{alg_vrnnaug_pred}, respectively. Note that the training process of VRNNaug in Algorithm~\ref{alg_vrnnaug_train} is conducted on the whole time series with length $T$. One could employ the aforementioned shingling technique to make data preprocessing and then use the mini-batch ELBO~\eqref{eq_elbo_unbiased} to train the model more efficiently. Besides, for line 4 in the prediction process of Algorithm~\ref{alg_vrnnaug_pred}, when $t=T+1$, we sample from previous state and output at $t=T$, which requires additional memory consumption; alternatively, one could start from cold by simply adopting the zero initialization. Finally, it is observed that the major difference of the for-loop in Algorithms~\ref{alg_vrnnaug_train} and~\ref{alg_vrnnaug_pred} is the estimation of the recurrent transformation $\bar{\mbf{y}}_{t-1}$. When training, we use the mixture of observation $\mbf{y}_{t-1}$ and prediction $\hat{\mbf{y}}_{t-1}$ in~\eqref{eq_hyb_y} and~\eqref{eq_ar_y}; contrarily, when predicting, without the access to observations, we only use the prediction $\hat{\mbf{y}}_{t-1}$ in~\eqref{eq_ar_y_hat}.

\begin{algorithm}[!t]
	\caption{Training of the proposed VRNNaug}
	\label{alg_vrnnaug_train}
	\begin{algorithmic}[1]
			\STATE {\bfseries Inputs:} Training time series data $\mathcal{D} = \{\mbf{u}_{1:T} \in \mathbb{R}^{T \times d_{\mbf{u}}}, \mbf{y}_{1:T} \in \mathbb{R}^{T \times d_{\mbf{y}}} \}$
			\STATE {\bfseries Output:} Sequence model $\mathcal{M}$ for future forecasting
			\REPEAT
			\FOR {$t=1:T$}
			\STATE Perform recurrent transformations $\bar{\mbf{z}}_{t-1}, \bar{\mbf{u}}_t, \bar{\mbf{y}}_{t-1}$ via~\eqref{eq_ar_z}-\eqref{eq_ar_y}
			\STATE Estimate mean $\bm{\mu}_{\mbf{z}_t}$ and variance $\bm{\nu}_{\mbf{z}_t}$ of $q(\mbf{z}_t|.)$ via~\eqref{eq_q_zt}
			\STATE Obtain mean $\bm{\mu}_{\mbf{y}_t}$ and variance $\bm{\nu}_{\mbf{y}_t}$ of $p(\mbf{y}_t|.)$ via~\eqref{eq_mu_nu_yt}
			\ENDFOR
			\STATE Calculate ELBO via~\eqref{eq_elbo}
			\STATE Update model parameters via stochastic optimizer
			\UNTIL{Optimizer has been terminated}
	\end{algorithmic} 
\end{algorithm}

\begin{algorithm}[!t]
	\caption{Prediction of the proposed VRNNaug}
	\label{alg_vrnnaug_pred}
	\begin{algorithmic}[1]
			\STATE {\bfseries Inputs:} Sequence model $\mathcal{M}$, forecast  horizon $F$, and input signals $\{\mbf{u}_t\}_{t=T+1}^{T+F}$
			\STATE {\bfseries Output:} $K$ prediction samples
			\FOR {$t=T+1:T+F$}
			\STATE Perform recurrent transformations $\bar{\mbf{z}}_{t-1}, \bar{\mbf{u}}_t, \bar{\mbf{y}}_{t-1}$ via~\eqref{eq_ar_z},~\eqref{eq_ar_u} and~\eqref{eq_ar_y_hat}
			\STATE Estimate mean $\bm{\mu}_{\mbf{z}_t}$ and variance $\bm{\nu}_{\mbf{z}_t}$ of $q(\mbf{z}_t|.)$ via~\eqref{eq_q_zt}
            \STATE Obtain mean $\bm{\mu}_{\mbf{y}_t}$ and variance $\bm{\nu}_{\mbf{y}_t}$ of $p(\mbf{y}_t|.)$ via~\eqref{eq_mu_nu_yt}
			\STATE Sample from $p(\mbf{y}_t|.)$ to produce $K$ prediction samples at time $t$
			\ENDFOR
	\end{algorithmic} 
\end{algorithm}

\subsection{Discussions} 
So far, we have described the proposed deep sequence model with known input signal $\mbf{u}_t$ at any time point, which includes additional covariate that may affect the output.  Note that our model is mainly designed for \textit{free} forecasting, i.e., multi-step ahead prediction. If one wants to conduct one-step ahead prediction, we could simply feed the ground truth $\mbf{y}_{t-1}$ into both the inference and generative networks in Fig.~\ref{fig_vrnnaug}. For time series without additional control signals, one could simply use the time features, for example, hour-of-day and day-of-week, as the additional $\mbf{u}_t$ for modeling.

Finally, the relation and difference of our model to existing literature should be highlighted. The most closely related works come from~\cite{bayer2014learning, fraccaro2018deep, li2021learning, gedon2020deep, salinas2020deepar}. In comparison to the similar VAE-type probabilistic sequence models, the proposed VRNNaug brings several improvements: (i) it adopts hybrid output in~\eqref{eq_hyb_y} during training for greatly alleviating the inconsistency issue; and (ii) it employs the generalized auto-regressive strategy in~\eqref{eq_bar_z}-\eqref{eq_bar_y} to form the recurrent inputs for better capturing long temporal dependency. The benefits brought by these improvements will be verified by the comparative study in section~\ref{sec_sysid} and the ablation study in section~\ref{sec_ablation}.

\section{Numerical experiments} \label{sec_exp}
This section first showcases the methodological characteristics of proposed VRNNaug model on two toy cases, followed by the extensive comparison against existing RNN assisted stochastic sequence models on eight system identification benchmarks and a real-world centrifugal compressor sensor data. 

We implement our model using PyTorch on a Linux workstation with TITAN RTX GPU. The detailed experimental configurations are provided in Appendix~\ref{app_exp_details}. As for the probabilistic forecasting provided by our model, we quantify the quality of predictive distribution by the $p$50- and $p$90-quantile losses~\cite{rangapuram2018deep, salinas2020deepar, chen2020probabilistic}. Generally, given the $i$-th time series output $\mbf{y}^i$ ($1 \le i \le d_{\mbf{y}}$) and its corresponding $\rho$-quantile prediction $\hat{\mbf{y}}^i_{\rho}$, the $\rho$-quantile loss is expressed as
\begin{align}
	\mathrm{QL}_{\rho}(\mbf{y}^i, \hat{\mbf{y}}^i_{\rho}) = 2 \times \frac{\sum_t P_{\rho}(y^i_t, \hat{y}^i_{\rho,t})}{\sum_t |y^i_t|}, \quad P_{\rho} = \left\{ 
	\begin{matrix}
		&\rho(y - \hat{y}_{\rho}), \quad &y > \hat{y}_{\rho}, \\
		&(1-\rho)(\hat{y}_{\rho} - y), \quad &\mathrm{otherwise}, \\
	\end{matrix}
	\right.
\end{align}
where the parameter $\rho \in [0,1]$. It is found that the $p50$ criterion using $\rho=0.5$ quantifies the relative error of prediction since $\mathrm{QL}_{0.5}(\mbf{y}^i, \hat{\mbf{y}}^i_{\rho}) = \sum_t |y^i_t - \hat{y}^i_{t}| / \sum_t |y^i_t|$ wherein $\hat{y}^i_{t}$ now is the medium of prediction samples at time $t$. As for the $p90$ criterion, it could represent the coverage of predictive distribution since if the observation falls below the 90\%-quantile prediction, we have a small $P_{\rho}=0.1(\hat{y}_{\rho}-y)$; otherwise, a large $P_{\rho}=0.9(y-\hat{y}_{\rho})$ should be used. Moreover, given desirable predictions, the predictive distribution with  uncertainty (i.e., a large variance) might not be favored by the $p90$ criterion because of the large difference $|\hat{y}_{\rho}-y|$ in $P_{\rho}$. Overall, lower $p50$ and $p90$ values indicate better quality of predictive distribution. Hence, in the following experiments, we use $p50 = \mathrm{QL}_{0.5}(.,.)$ to quantify the accuracy of prediction medium, and $p90 = \mathrm{QL}_{0.9}(.,.)$ to represent the coverage of predictive distribution.

\subsection{Toy cases}
We below study the proposed VRNNaug model on two toy cases (the linear Gaussian system and the heteroscedastic \texttt{motorcycle} case) with different features.

\textbf{Linear Gaussian system.} We first consider a one-dimensional linear Gaussian system~\cite{gedon2020deep} expressed as
\begin{align} \label{eq_lgssm}
	\mbf{h}_{t+1} =& 
	\left[ 
	\begin{matrix}
		0.7 & 0.8 \\
		0 & 0.1 \\
	\end{matrix}
	\right] \mbf{h}_t +
	\left[ 
	\begin{matrix}
		-1 \\
		0.1 \\
	\end{matrix} 
	\right] u_t + \bm{\epsilon}_h, \\
	y_t =& 
	\left[ 
	\begin{matrix}
		1 & 0 \\
	\end{matrix} 
	\right] \mbf{h}_t + \epsilon,
\end{align}
where $\mbf{h} \in \mathbb{R}^2$ is a two-dimensional hidden state, $\bm{\epsilon}_h \sim \mathcal{N}(\mbf{0}, 0.5\mbf{I})$ is the process noise, and $\epsilon \sim \mathcal{N}(0, 1)$ is the measurement noise. We use this dynamic model to sequentially generate 2000 training points, 2000 validation points and 5000 test points. Note that when generating training and validation signals, an excitation input signal $u_t \in [-2.5, 2.5]$ polluted with uniform random noise is employed for the linear system. Differently, the input signal for the test data writes as
\begin{align*}
	u_t = \sin\left(\frac{2t\pi}{10} \right) + \sin\left(\frac{2t\pi}{25} \right).
\end{align*}

\begin{figure}[t!]
	\centering
	\includegraphics[width=0.6\textwidth]{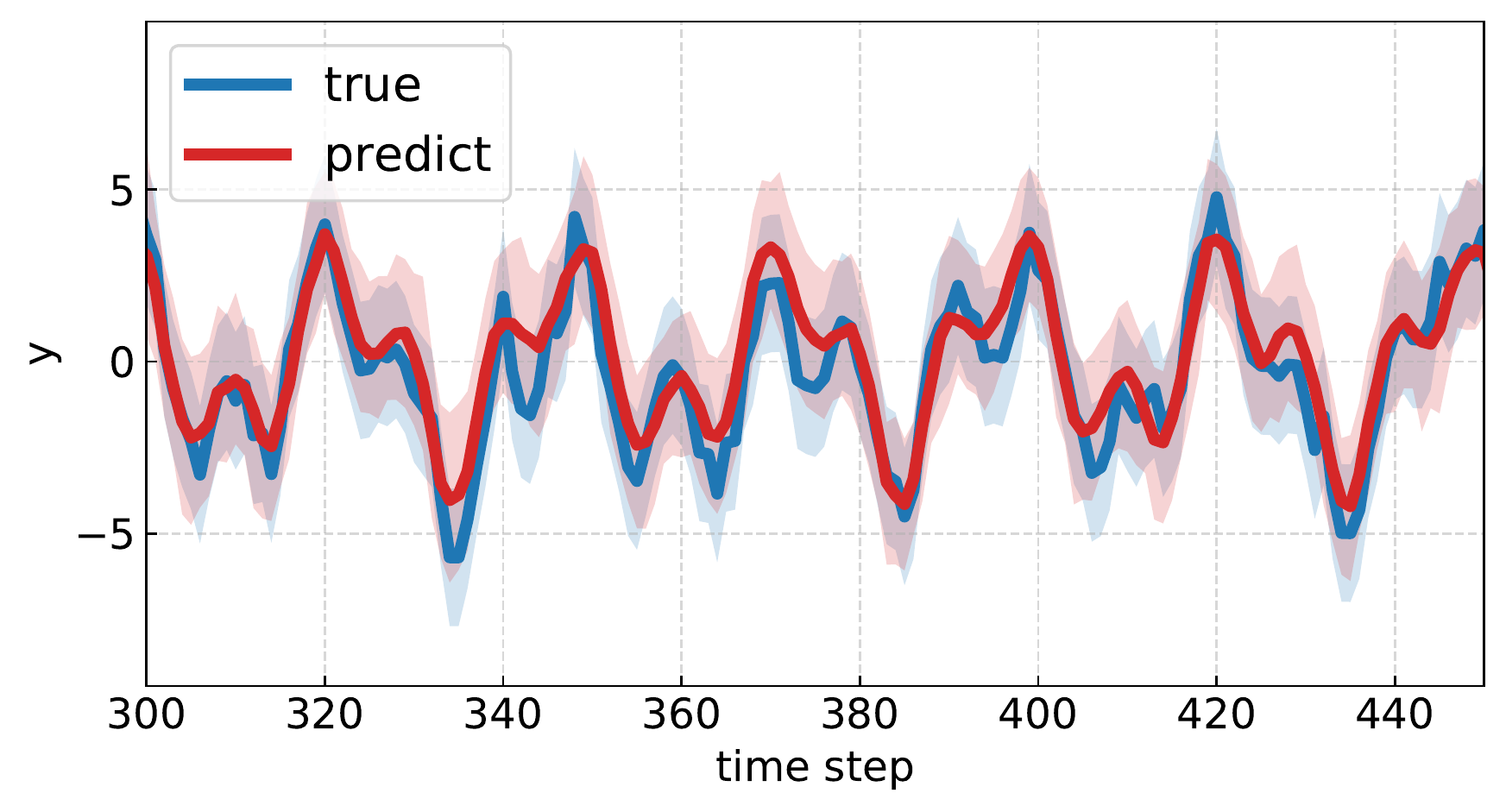}
	\caption{The predictive distribution of VRNNaug on the linear Gaussian system. The red curve represents 50-quantile prediction, while the associated shaded region indicates the interval bounded by 95-quantile and 5-quantile predictions. The blue curve is the true observation with the shaded region representing 95\% confidence interval.}
	\label{fig_lgssm}
\end{figure}

The predictive distributions of the proposed VRNNaug together with the true observations are depicted in Fig.~\ref{fig_lgssm}. It is observed that for this linear system, the flexible and nonlinear VRNNaug model successfully captures the underlying linear dynamics. This linear case has also been used in~\cite{gedon2020deep} to test another deep SSM model, named STORN~\cite{bayer2014learning}, and the gray-box model SSEST~\cite{lennart1999system} with two latent states. Though successfully capturing the dynamics, they conservatively estimate the uncertainty of prediction. Our model however is found to well estimate the uncertainty in Fig.~\ref{fig_lgssm}.

\textbf{Heteroscedastic motorcycle data.} We next study a more complicated \texttt{motorcycle} time series dataset~\cite{silverman1985some} composed of 133 points with time input and acceleration output. The \texttt{motorcycle} data takes from an experiment on the efficacy of crash helmets, and yields multiple acceleration observations at some time points. In comparison to the previous linear Gaussian system with homoscedastic noise, this nonlinear dataset contains more challenging \textit{time-varying} noise.

\begin{figure}[t!]
	\centering
	\includegraphics[width=0.6\textwidth]{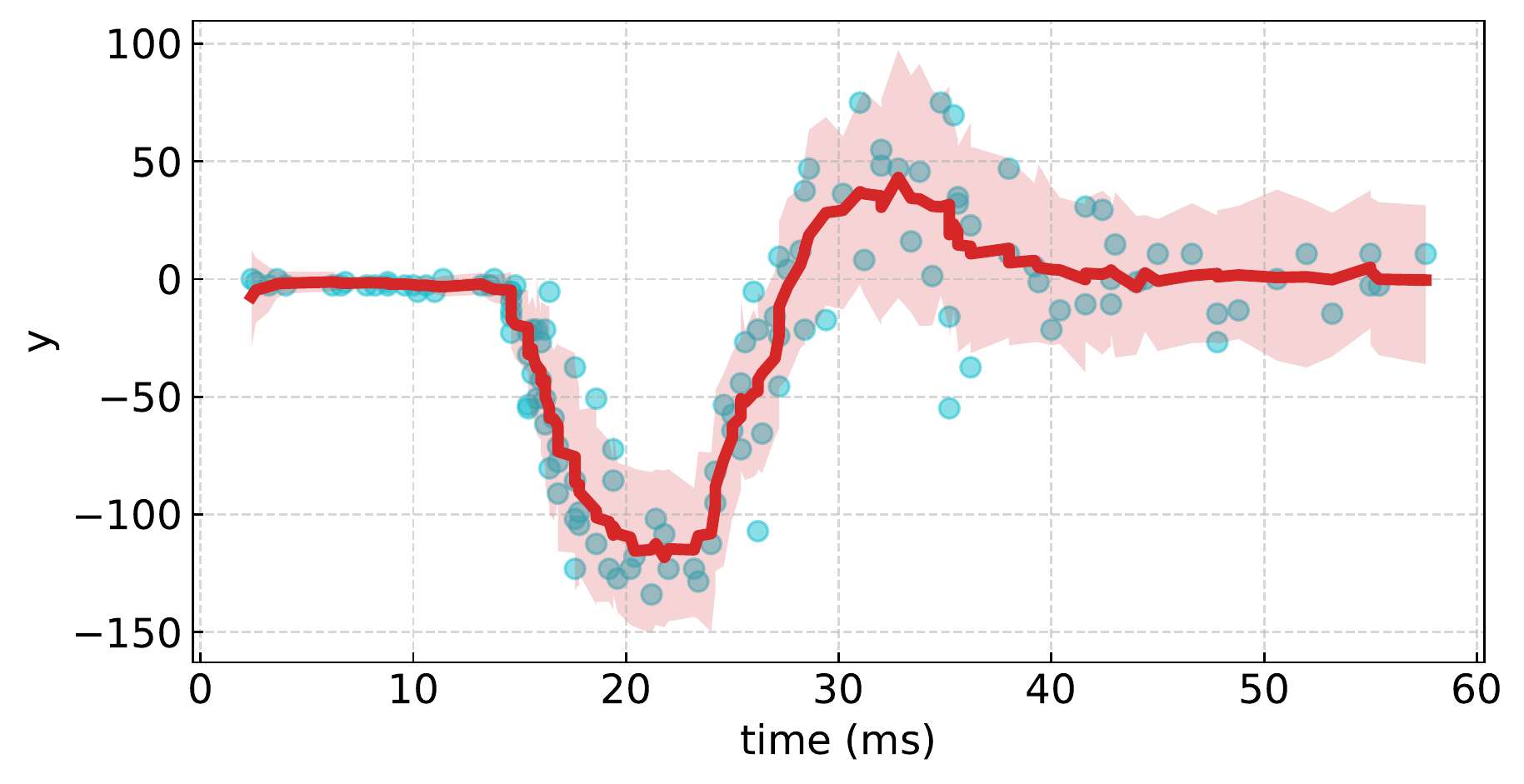}
	\caption{The predictive distribution of VRNNaug on the heteroscedastic \texttt{motorcycle} dataset. The red curve represents 50-quantile prediction, while the associated shaded region indicates the interval bounded by 95-quantile and 5-quantile predictions. The circles represent 133 data points.}
	\label{fig_motor}
\end{figure}

We seek to verify the capability of VRNNaug for capturing nonlinear dynamic patterns with heteroscedastic noise on this case. We here use all the 133 data points for training, validation and testing, and extract the time feature as input signal for our VRNNaug model. As depicted in Fig.~\ref{fig_motor}, the proposed VRNNaug captures the non-linear dynamics of underlying acceleration curve, for example, the constant behavior before $t=15$ ms and the step behavior around $t=15$ ms. More interestingly, it well describes the time-varying noise, for example, the pretty small perturbation before $t=15$ ms and the large and time-varying noise thereafter.

\subsection{System identification benchmarks} \label{sec_sysid}
This section verifies the performance of proposed deep sequence model on eight popular system identification benchmarks from dynamics systems like hydraulic actuator, electric motors, hair dryer, furnace, tank, cascaded tanks system~\cite{schoukens2016cascaded}, F-16 aircraft ground vibration test~\cite{noel2017f}, and wiener-hammerstein process noise system~\cite{schoukens2016wiener}.\footnote{The first five datasets are available at \url{https://homes.esat.kuleuven.be/~smc/daisy/daisydata.html} and the last three datasets are collected at \url{http://nonlinearbenchmark.org/index.html}.} Table~\ref{tab_sys_iden} summarizes the sequence length and dimensions of the considered time series datasets. It is found that all these datasets except \texttt{f16gvt} have a single input signal, and the particular \texttt{tank} dataset has two outputs. The goal of sequence modeling is to infer the underlying high-dimensional latent state and the relevant nonlinear dynamics from data, and use them for future forecasting.

\begin{table}
	\caption{Characteristics of the system identification benchmarks.} 
	\label{tab_sys_iden}
	\centering
	\begin{tabular}{lrrr}
		\hline
		dataset &$T$ &$d_{\mbf{u}}$ &$d_{\mbf{y}}$ \\
		\hline
		actuator &1024 & 1 & 1 \\
		cascaded tank (ctank) &2028 & 1 & 1 \\
		drive &500 & 1 & 1 \\
		dryer &1000 & 1 & 1 \\
		f16gvt &32768 & 2 & 1 \\
		gas furnace (furnace) &296 & 1 & 1 \\
		tank &2500 & 1 & 2 \\
		wiener hammerstein (hammerstein) &32768 & 1 & 1 \\
		\hline
	\end{tabular}
\end{table}

The performance of VRNNaug is compared against state-of-the-art RNN assisted deep stochastic SSM models, including (i) the linear SSM model with parameters learned through neural networks (Deep-LSSM)~\cite{rangapuram2018deep}; (ii) the VAE-RNN model with the encoder built upon RNN~\cite{fraccaro2018deep}; (iii) the stochastic RNN (STORN) following the encoder-decoder structure~\cite{bayer2014learning}; (iv) the deep auto-regressive model (Deep-AR)~\cite{salinas2020deepar}; and finally, (v) the deep SSM for probabilistic forecasting, donated as DSSMF~\cite{li2021learning}.\footnote{The DSSMF in~\cite{li2021learning} additionally adopts an automatic relevance determination network to identify the importance of dimensions of the input signal $\mbf{u}_t$, which is not implemented in our comparison.} Tables~\ref{tab_sys_iden_p50} and~\ref{tab_sys_iden_p90} report the $p50$ and $p90$ results of various deep stochastic SSM models over ten runs, respectively. Note that the best and second-best results on the eight benchmarks are marked as gray and light gray, respectively. 

\begin{sidewaystable}
	\caption{The $p$50 results of various deep stochastic SSM models on the eight system identification benchmarks. The best and second-best results on each dataset are marked as gray and light gray, respectively.} 
	\label{tab_sys_iden_p50}
	\centering
		\begin{tabular}{lrrrrrr}
			\hline
			Datasets &Deep-LSSM &VAE-RNN &STORN &Deep-AR  & DSSMF & VRNNaug\\
			\hline
			\texttt{actuator}	&0.3866$_{\pm0.0225}$	&0.3529$_{\pm0.0263}$	&\cellcolor{mygray}0.3503$_{\pm0.0104}$
			&0.3839$_{\pm0.0407}$ &0.3737$_{\pm0.0536}$	&\cellcolor{lightgray}0.3241$_{\pm0.0265}$	\\ 
			\texttt{ctank}	&0.2777$_{\pm0.0469}$	&0.1558$_{\pm0.0174}$	&0.1065$_{\pm0.0092}$
			&\cellcolor{mygray}0.0965$_{\pm0.0082}$ &0.1028$_{\pm0.0333}$	&\cellcolor{lightgray}0.0715$_{\pm0.0192}$	\\
			\texttt{drive}	&\cellcolor{mygray}0.3157$_{\pm0.0359}$	&0.4925$_{\pm0.0057}$	&0.3792$_{\pm0.0599}$
			&0.5083$_{\pm0.0072}$ &0.5056$_{\pm0.0119}$	&\cellcolor{lightgray}0.2098$_{\pm0.0186}$	\\
			\texttt{dryer}	&0.0201$_{\pm0.0011}$	&0.0203$_{\pm0.0013}$	&0.0162$_{\pm0.0010}$
			&0.0179$_{\pm0.0016}$ &\cellcolor{lightgray}0.0153$_{\pm0.0025}$	&\cellcolor{mygray}0.0158$_{\pm0.0017}$	\\
			\texttt{f16gvt}	&0.7108$_{\pm0.1586}$	&\cellcolor{lightgray}0.0723$_{\pm0.0072}$	&\cellcolor{mygray}0.0735$_{\pm0.0044}$
			&0.2973$_{\pm0.0111}$ &0.3021$_{\pm0.0124}$	&0.1540$_{\pm0.0140}$	\\
			\texttt{furnace}	&0.0238$_{\pm0.0021}$	&0.0436$_{\pm0.0147}$	&\cellcolor{lightgray}0.0225$_{\pm0.0006}$
			&0.0231$_{\pm0.0005}$ &0.0242$_{\pm0.0009}$	&\cellcolor{mygray}0.0226$_{\pm0.0008}$	\\
			\texttt{tank} ($y_1$)	&0.1005$_{\pm0.0203}$	&0.1885$_{\pm0.0155}$	&0.0785$_{\pm0.0212}$
			&\cellcolor{mygray}0.0539$_{\pm0.0169}$ &\cellcolor{lightgray}0.0529$_{\pm0.0072}$	&0.0591$_{\pm0.0154}$	\\
			\texttt{tank} ($y_2$)	&0.0597$_{\pm0.0229}$	&0.0615$_{\pm0.0104}$	&0.0559$_{\pm0.0083}$
			&\cellcolor{lightgray}0.0414$_{\pm0.0085}$ &0.0437$_{\pm0.0042}$	&\cellcolor{mygray}0.0429$_{\pm0.0063}$	\\
			\texttt{hammerstein} &NA	&0.9400$_{\pm0.1816}$	&0.7364$_{\pm0.0526}$
			&\cellcolor{mygray}0.4676$_{\pm0.0582}$ &0.5088$_{\pm0.0972}$	&\cellcolor{lightgray}0.2261$_{\pm0.0777}$	\\
			\hline
		\end{tabular}
\end{sidewaystable}

\begin{sidewaystable}
	\caption{The $p$90 results of various deep stochastic SSM models on the eight system identification benchmarks. The best and second-best results on each dataset are marked as gray and light gray, respectively.} 
	\label{tab_sys_iden_p90}
	\centering
		\begin{tabular}{lrrrrrr}
			\hline
			Datasets &Deep-LSSM &VAE-RNN &STORN &Deep-AR  & DSSMF & VRNNaug\\
			\hline
			\texttt{actuator}	&0.2059$_{\pm0.0226}$	&0.2302$_{\pm0.0280}$	&\cellcolor{mygray}0.1894$_{\pm0.0124}$
			&0.1965$_{\pm0.0214}$ &0.1925$_{\pm0.0172}$	&\cellcolor{lightgray}0.1776$_{\pm0.0158}$	\\ 
			\texttt{ctank}	&0.2209$_{\pm0.0826}$	&0.0979$_{\pm0.0111}$	&0.0656$_{\pm0.0093}$
			&0.0758$_{\pm0.0179}$ &\cellcolor{mygray}0.0611$_{\pm0.0192}$	&\cellcolor{lightgray}0.0358$_{\pm0.0138}$	\\ 
			\texttt{drive}	&\cellcolor{mygray}0.1610$_{\pm0.0164}$	&0.1966$_{\pm0.0039}$	&0.1737$_{\pm0.0276}$
			&0.2173$_{\pm0.0101}$ &0.2245$_{\pm0.0118}$	&\cellcolor{lightgray}0.1029$_{\pm0.0058}$	\\
			\texttt{dryer}	&0.0079$_{\pm0.0010}$	&0.0093$_{\pm0.0015}$	&\cellcolor{lightgray}0.0060$_{\pm0.0009}$
			&0.0082$_{\pm0.0012}$ &\cellcolor{mygray}0.0062$_{\pm0.0015}$	&0.0067$_{\pm0.0012}$	\\
			\texttt{f16gvt}	&0.3180$_{\pm0.0664}$	&\cellcolor{lightgray}0.0342$_{\pm0.0026}$	&\cellcolor{mygray}0.0353$_{\pm0.0018}$
			&0.1310$_{\pm0.0053}$ &0.1356$_{\pm0.0059}$	&0.0733$_{\pm0.0065}$	\\
			\texttt{furnace}	&0.0210$_{\pm0.0009}$	&\cellcolor{lightgray}0.0176$_{\pm0.0013}$	&0.0221$_{\pm0.0007}$
			&0.0229$_{\pm0.0012}$ &0.0222$_{\pm0.0010}$	&\cellcolor{mygray}0.0197$_{\pm0.0009}$	\\
			\texttt{tank} ($y_1$)	&\cellcolor{mygray}0.0386$_{\pm0.0089}$	&0.1219$_{\pm0.0127}$	&0.0605$_{\pm0.0182}$
			&0.0592$_{\pm0.0281}$ &0.0566$_{\pm0.0115}$	&\cellcolor{lightgray}0.0318$_{\pm0.0159}$	\\
			\texttt{tank} ($y_2$)	&0.0287$_{\pm0.0160}$	&\cellcolor{mygray}0.0286$_{\pm0.0084}$	&0.0344$_{\pm0.0109}$
			&0.0463$_{\pm0.0136}$ &0.0504$_{\pm0.0072}$	&\cellcolor{lightgray}0.0285$_{\pm0.0091}$	\\
			\texttt{hammerstein} &NA	&0.3992$_{\pm0.0603}$	&0.3689$_{\pm0.0288}$
			&\cellcolor{mygray}0.2024$_{\pm0.0313}$ &0.2122$_{\pm0.0339}$	&\cellcolor{lightgray}0.1443$_{\pm0.0212}$	\\
			\hline
		\end{tabular}
\end{sidewaystable}

It is first observed that due to the \textit{linear} SSM framework, the Deep-LSSM is hard to learn the nonlinear dynamic behaviors from sequence data, even though it attempts to learn the local, time-varying coefficients. For instance, though the Deep-LSSM forecasts the rough profile on the \texttt{ctank} dataset, its prediction however leaves far away from the ground truth in comparison to the others, which is indicated by the poor $p50$ results in Table~\ref{tab_sys_iden_p50}. Furthermore, it fails on the large and complicated \texttt{hammerstein} dataset. Therefore, in comparison to other competitors, the Deep-LSSM showcases the worst performance in general in terms of both $p50$ and $p90$. Despite the linear model structure, the undesirable performance of Deep-LSSM may be attributed to the type-I maximum likelihood framework wherein the latent variables are not integrated out.

\begin{figure}[t!]
	\centering
	\includegraphics[width=0.8\textwidth]{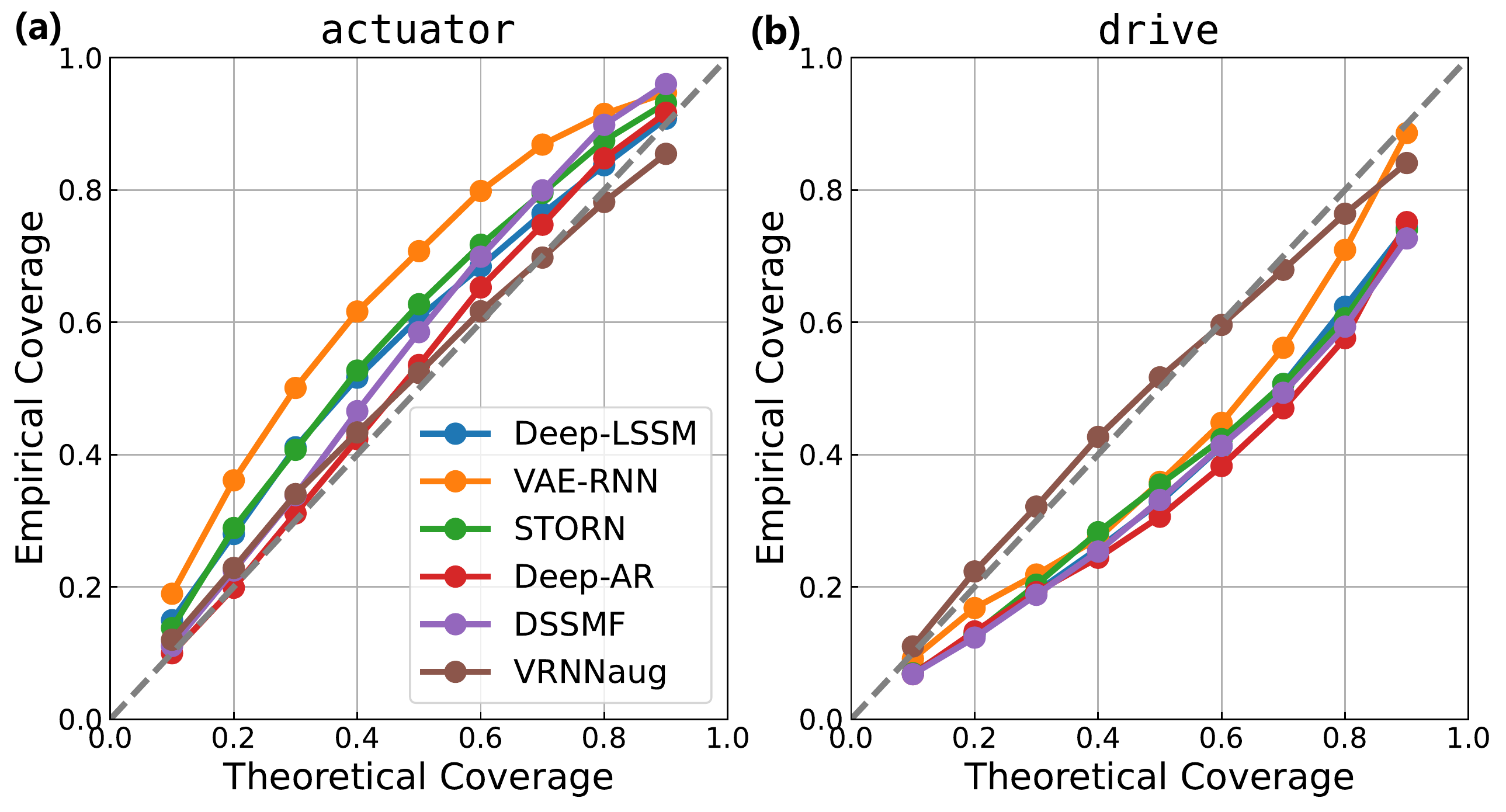}
	\caption{Coverage plots of predictive distributions of various probabilistic sequence models on the (a) \texttt{actuator} and (b) \texttt{drive} datasets.}
	\label{fig_ablation_drive_ecp}
\end{figure}

As for the VAE-type deep SSM models, the more complicated STORN is found to outperform the simple VAE-RNN on most cases in terms of both $p50$ and $p90$. This is attributed to the description of the recurrency in previous outputs $\mbf{d}_t = \mathrm{RNN}(\mbf{y}_{1:t})$ when determining the posterior $q(\mbf{z}_t|.)$. But as has been discussed, the usage of current output $\mbf{y}_t$ at time $t$ raises the inconsistency issue: when forecasting time series through generative networks, we have no access to current output and then have to sample $\hat{\mbf{z}}_t$ from prior. This thus makes STORN's $p50$ results in Table~\ref{tab_sys_iden_p90} slightly worse than that of Deep-AR and DSSMF which use the auto-regressive strategy to alleviate the issue. Besides, the sampling from prior when generating time series data makes VAE-RNN and STORN tend to provide relatively large uncertainty estimation, which will be illustrated in Fig.~\ref{fig_ablation_drive_ecp} as well as the \texttt{compressor} case in next section.

As for the last three deep SSM models, all of them employ sort of auto-regressive strategy to avoid the inconsistency issue. In comparison to Deep-AR and DSSMF, the proposed VRNNaug adopts (i) the hybrid output to further alleviate the gap between training and predicting; and (ii) the generalized auto-regressive strategy for extracting long term historical patterns. As a result, it performs the best on these system identification benchmarks in terms of both $p50$ and $p90$.\footnote{The VRNNaug forecasting in future time period on the eight system identification benchmarks is respectively illustrated in Appendix~\ref{app_benchmarks}.} Besides, it is found that akin to Deep-LSSM, the Deep-AR also trains the model in the simple type-I maximum-likelihood framework, thus making it perform slightly worse than the competitor DSSMF derived through variational inference, especially in terms of the $p90$ criterion.

Finally, to fully represent the quality of prediction distribution of the above probabilistic sequence models, Fig.~\ref{fig_ablation_drive_ecp} adopts the empirical coverage probability (ECP) criterion used in~\cite{yanchenko2020stanza} to compare their performance on the \texttt{actuator} and \texttt{drive} datasets. It is found that the proposed VRNNaug achieves good empirical coverage on the two datasets. As for VAE-RNN, its ECP is higher than the theoretical coverage on the \texttt{actuator} dataset, which implies conservative uncertainty that has also been identified by the worst $p90$ value in Table~\ref{tab_sys_iden_p90}. As for the challenging \texttt{drive} dataset, since the predictions of all the models except VRNNaug are usually far away from the observations,\footnote{For example, the VAE-RNN fails on the \texttt{drive} dataset by providing almost the constant predictions.} indicated by the poor $p50$ results in Table~\ref{tab_sys_iden_p50}, their ECP results in Fig.~\ref{fig_ablation_drive_ecp}(b) are not good. The exception of good 90\% ECP of VAE-RNN in Fig.~\ref{fig_ablation_drive_ecp}(b) is solely attributed to the large uncertainty. 

\subsection{Ablation study} \label{sec_ablation}
With the good performance reported in Tables~\ref{tab_sys_iden_p50} and~\ref{tab_sys_iden_p90}, we here perform ablation study of the proposed VRNNaug model on the \texttt{drive} dataset. The ablation study includes two additional VRNNaug variants in order to reveal individual effects of components: (i) the VRNNaug-v1 which resembles Deep-AR by using only the auto-regressive strategy via $\mbf{y}_{t-1}$; and (ii) the VRNNaug-v2 which uses the generalized auto-regressive strategy in~\eqref{eq_ar_z} and~\eqref{eq_ar_u}, but simply uses $\mbf{y}_{t-1}$ instead of the hybrid $\dot{\mbf{y}}_{t-1}$ in~\eqref{eq_ar_y} to determine $\bar{\mbf{y}}_{t-1}$. Differently, the complete VRNNaug model adopts both the previous hybrid outputs and the generalized auto-regressive strategy to alleviate the inconsistency issue and improve the model capability.

\begin{figure}[t!]
	\centering
	\includegraphics[width=0.5\textwidth]{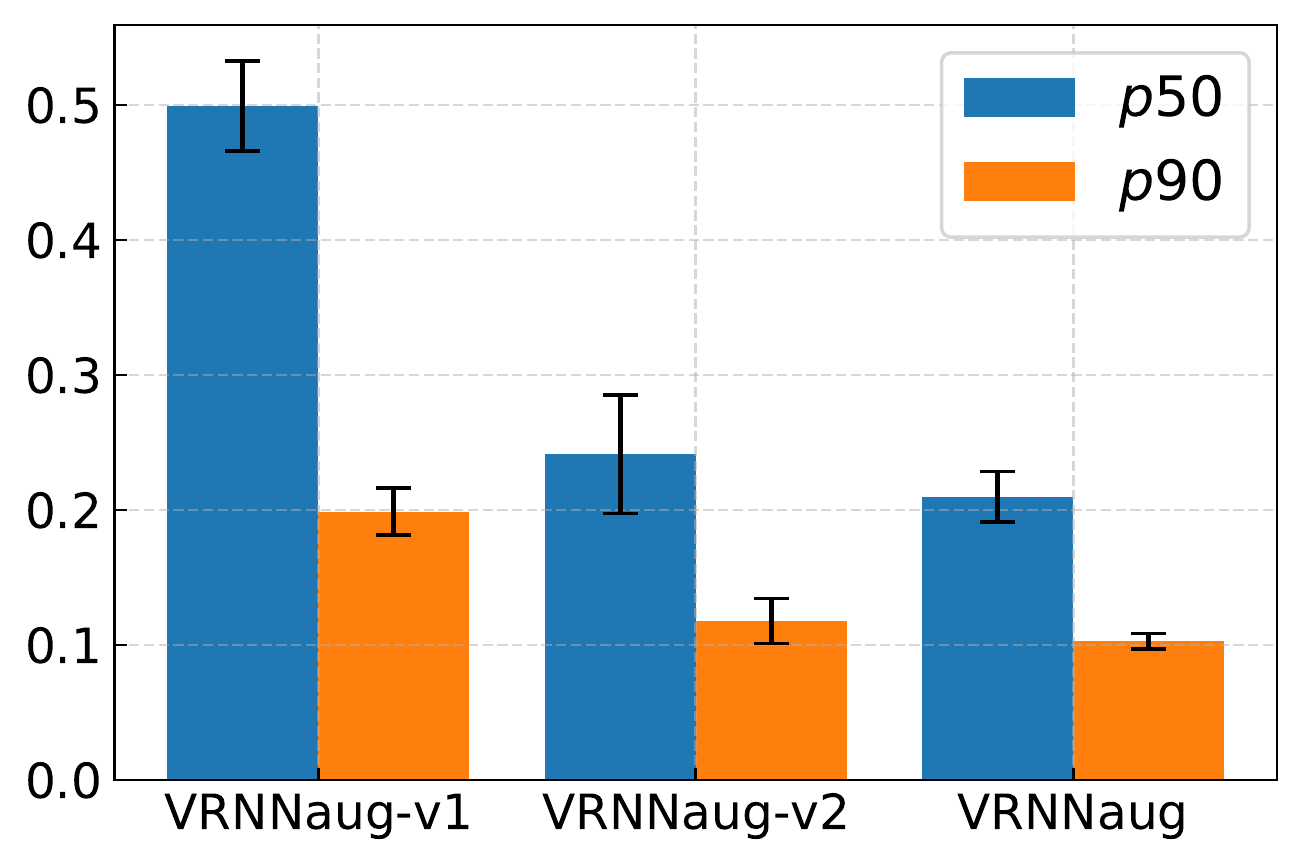}
	\caption{Ablation study of the proposed VRNNaug model on the \texttt{drive} dataset.}
	\label{fig_ablation_driver}
\end{figure}

Fig.~\ref{fig_ablation_driver} depicts the comparative results of three VRNNaug models in terms of both the $p50$ and $p90$ criteria on the \texttt{drive} dataset, and has the following findings.
\begin{itemize}
	\item By solely using the auto-regressive strategy akin to Deep-AR and DSSMF, the VRNNaug-v1 is incapable of capturing the underlying nonlinear dynamical behavior on the challenging \texttt{drive} dataset. Contrarily, the generalized auto-regressive strategy summarizes all the long term historical patterns of not only $\mbf{y}$ but also $\mbf{z}$ and $\mbf{u}$. Consequently, both VRNNaug-v2 and VRNNaug significantly outperform the basic VRNNaug-v1;
	\item The hybrid output $\dot{\mbf{y}}_{t-1}$ used in training helps further fill the gap between inference and generation, since now we always consider the prediction samples in the two phases. Hence, it brings considerable improvements for VRNNaug over VRNNaug-v2.
\end{itemize}

\subsection{Axis displacement forecasting of centrifugal compressor}
\begin{figure}[t!]
	\centering
	\includegraphics[width=0.7\textwidth]{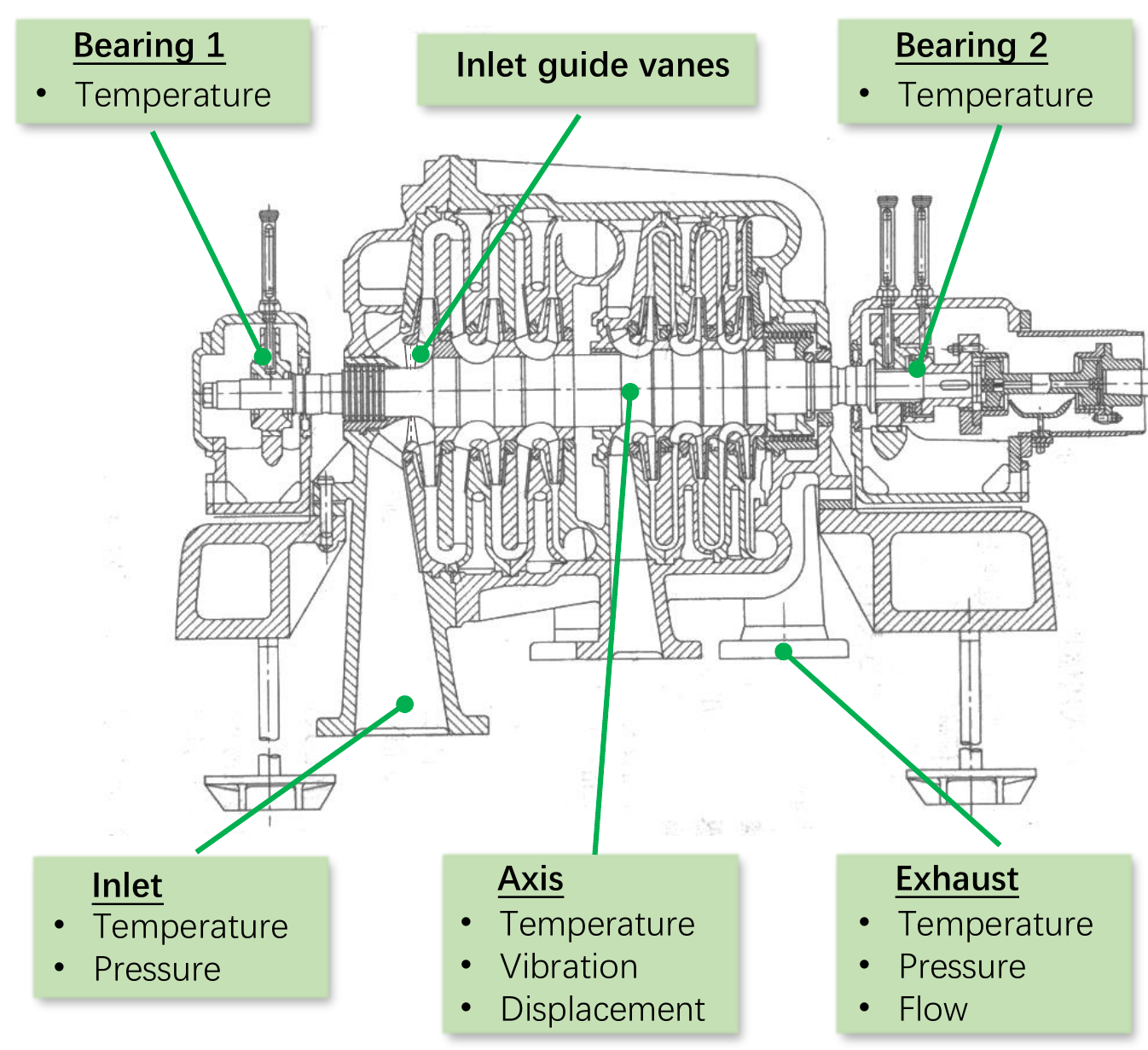}
	\caption{Illustration of sensors layout for collecting the \texttt{compressor} time series data.}
	\label{fig_compressor_sensor}
\end{figure}

This section verifies the performance of the proposed VRNNaug on an 11-variable time series data collected from a real-world centrifugal compressor with the rotating speed of 5556 RPM~\cite{xu2016bayesian}. As illustrated in Fig.~\ref{fig_compressor_sensor}, this time series data was recorded every one hour from March to October, 2013, by placing sensors at the inlet, the inlet guide vanes, the exhaust, the bearing, and the axis of centrifugal compressor, thus resulting in 3710 raw data points for each variable. 

As for this real-world time series task, we perform data preprocessing before applying deep stochastic SSM models on it for forecasting. These 11-variable time series data is collected manually from sensors, thus may containing human errors, outliers, noise and missing values. Hence, we first perform outlier analysis to identify and remove the outliers in each raw time series variable via the box-plot method, thus resulting in 3524 data points. Secondly, in order to eliminate the effect of magnitudes within multivariate time series, we normalize all the variables into the same range $[0,1]$. Thirdly, we conduct discrete wavelet packet transform (DWPT) analysis by the Daubechies wavelet of eight-order~\cite{daubechies1992ten} as well as the Bayesian thresholding technique in order to denoise our time series data. 

By processing and denoising the raw data, we finally obtain the \texttt{compressor} time series containing 3524 data points. Following the experimental configurations elaborated in Appendix~\ref{app_exp_details}, our aim is to predict the axis displacement at the last 10\% time points, given the remaining ten variables as covariates.

\begin{table}
	\caption{The $p50$ and $p90$ results of various deep stochastic SSM models on the real-world \texttt{compressor} dataset. Note that the best and second-best results on each dataset are marked as gray and light gray, respectively.} 
	\label{tab_compressor}
	\centering
    \begin{tabular}{lrr}
			\hline
			Method &$p50$ &$p90$ \\
			\hline
			Deep-LSSM	&0.0597$_{\pm0.0119}$	&0.0428$_{\pm0.0106}$ \\
			VAE-RNN &0.0563$_{\pm0.0093}$ &\cellcolor{mygray}0.0259$_{\pm0.0040}$ \\
			STORN &\cellcolor{mygray}0.0540$_{\pm0.0067}$ &0.0270$_{\pm0.0055}$ \\
			Deep-AR &0.0674$_{\pm0.0294}$ &0.0531$_{\pm0.0371}$ \\
			DSSMF &0.0600$_{\pm0.0188}$ &0.0581$_{\pm0.0351}$ \\
			VRNNaug &\cellcolor{lightgray}0.0488$_{\pm0.0050}$ &\cellcolor{lightgray}0.0255$_{\pm0.0068}$ \\
			\hline
	\end{tabular}
\end{table}

\begin{figure*}[t!]
	\centering
	\begin{subfigure}
		\centering
		\includegraphics[width=1.\textwidth]{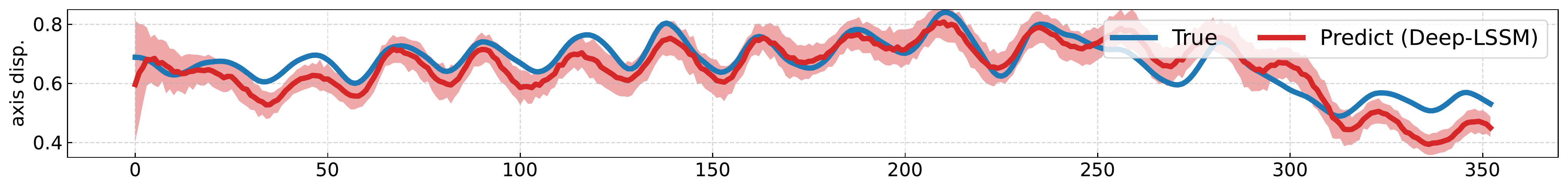}
	\end{subfigure}%
	\begin{subfigure}
		\centering
		\includegraphics[width=1.\textwidth]{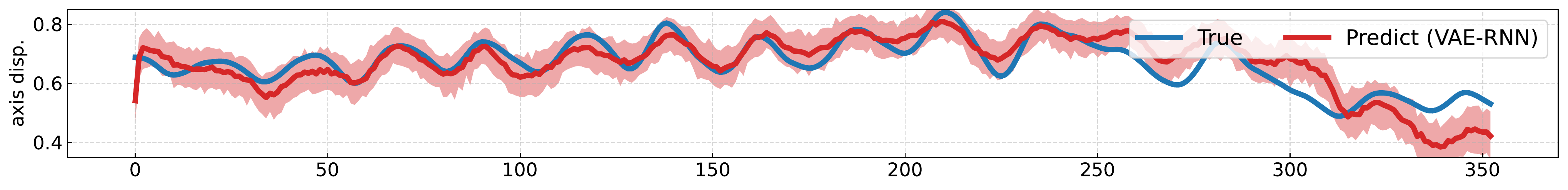}
	\end{subfigure}
	\begin{subfigure}
		\centering
		\includegraphics[width=1.\textwidth]{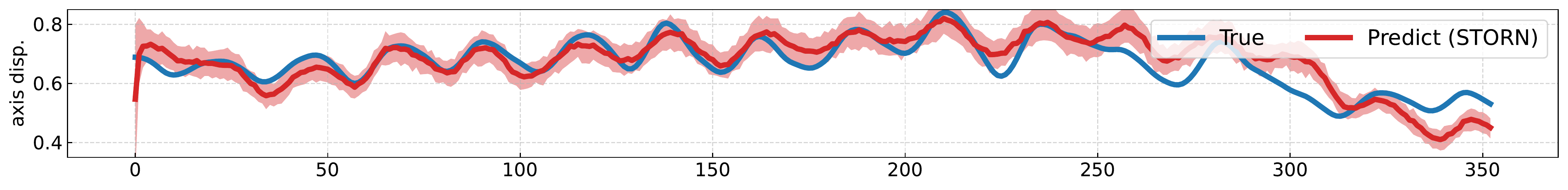}
	\end{subfigure} 
	\begin{subfigure}
		\centering
		\includegraphics[width=1.\textwidth]{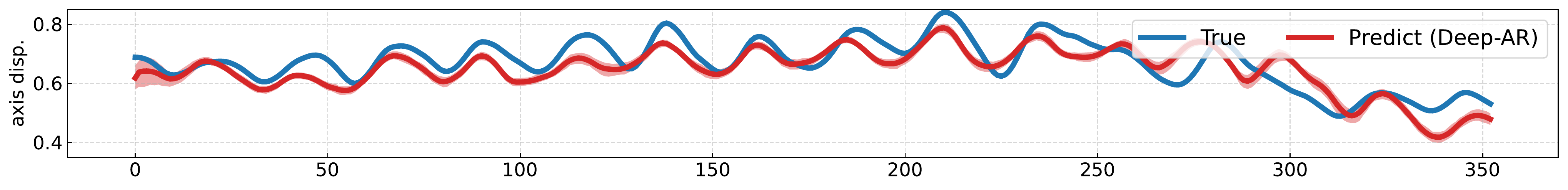}
	\end{subfigure}  
	\begin{subfigure}
		\centering
		\includegraphics[width=1.\textwidth]{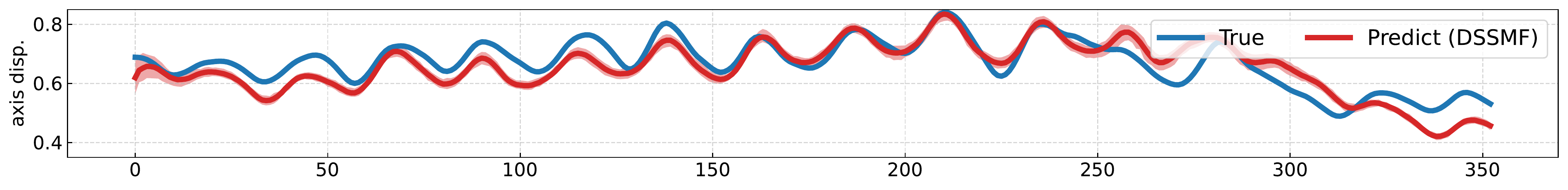}
	\end{subfigure} 
	\begin{subfigure}
		\centering
		\includegraphics[width=1.\textwidth]{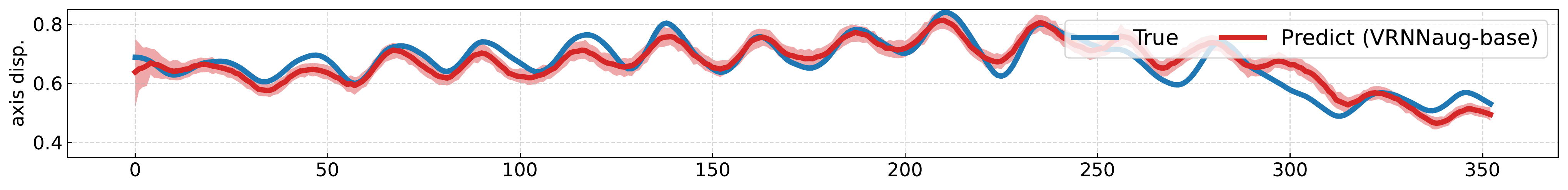}
	\end{subfigure} 
	\caption{The future axis displacement of centrifugal compressor forecasted by various deep SSM models. The red curve represents 50-quantile prediction, while the associated shaded region indicates the interval bounded by 95-quantile and 5-quantile predictions.}
	\label{fig_compressor_pred} 
\end{figure*}

Table~\ref{tab_compressor} reports the forecasting results of VRNNaug against that of other deep stochastic SSM models on the real-world \texttt{compressor} dataset. The superiority of VRNNaug over other competitors has again been observed in terms of both the $p50$ and $p90$ criteria for forecasting the future axis displacement. Moreover, Fig.~\ref{fig_compressor_pred} illustrates the forecasting of these deep stochastic SSM models for the future axis displacement. It is observed that the proposed VRNNaug provides more accurate predictions. Besides, the uncertainty of Deep-AR and DSSMF is relatively small, thus indicated by the poor $p90$ results in Table~\ref{tab_compressor}. Contrarily, the uncertainty of VAE-RNN and STORN is relatively large, which may be attributed to the sampling from prior $p(\mbf{z}_t|.)$ rather than the informative posterior $q(\mbf{z}_t|.)$ in the generative network.\footnote{Note that in comparison to the small uncertainty of Deep-AR and DSSMF, this conservative uncertainty is favored by the $p90$ criterion since it might cover most observations on the \texttt{compressor} case.} The proposed VRNNaug achieves a trade-off between the two extreme cases, thus providing desirable estimation of uncertainty, which is revealed by the best $p90$ results.

\section{Conclusions} \label{sec_remarks}
In this paper, we present a variational RNN model on the augmented recurrent space for probabilistic time series forecasting. When using variational inference to derive the ELBO as objective for model training, we figure out the issue of inconsistency that would deteriorate the model performance. Hence, we propose (i) feeding the \textit{hybrid} output as input at next time step, and (ii) presenting the \textit{generalized} auto-regressive strategy in order to better extract temporal patterns. Through extensive experiments on various time series tasks against existing deep stochastic SSM models, we have showcased the superiority of proposed VRNNaug model by quantifying predictive distribution. Future work would attempt to extend the proposed model to challenging scenarios in dynamic systems with for example missing outputs, irregular time points, and knowledge transfer across multivariate time series.

\section*{Acknowledgments}
This work was supported by the National Natural Science Foundation of China (52005074), and the Fundamental Research Funds for the Central Universities (DUT19RC(3)070). Besides, it was partially supported by the Research and Innovation in Science and Technology Major Project of Liaoning Province (2019JH1-10100024), and the MIIT Marine Welfare Project (Z135060009002).

\begin{appendices}
\section{Acronyms} \label{app_acronyms}
$
\begin{array}{ll}
	\mbox{ARIMA} & \mbox{Auto-Regressive Integrated Moving Average} \\
	\mbox{CFP} & \mbox{Empirical Coverage Probability} \\
	\mbox{Deep-AR} & \mbox{Deep Auto-regressive Model~\cite{salinas2020deepar}} \\
	\mbox{Deep-LSSM} & \mbox{Deep Linear State Space Model~\cite{rangapuram2018deep}} \\
	\mbox{DSSMF} & \mbox{Deep State Space Model for Probabilistic Forecasting~\cite{li2021learning}} \\
	\mbox{DWPT} & \mbox{Discrete Wavelet Packet Transform} \\
	\mbox{ELBO} & \mbox{Evidence Lower Bound} \\
	\mbox{GRU} & \mbox{Gated Recurrent Unit} \\
	\mbox{KL} & \mbox{Kullback-Leibler} \\
	\mbox{LSTM} & \mbox{Long Short Term Memory} \\
	\mbox{MLP} & \mbox{Multi-layer Perceptron} \\
	\mbox{NLP} & \mbox{Natural Language Processing} \\
	\mbox{PHM} & \mbox{Prognostic and Health Management} \\
	\mbox{RNN} & \mbox{Recurrent Neural Networks} \\
	\mbox{SSM} & \mbox{State Space Model} \\
	\mbox{STORN} & \mbox{Stochastic RNN~\cite{bayer2014learning}} \\
	\mbox{VAE} & \mbox{Variational Autoencoder~\cite{fraccaro2018deep}} \\
	\mbox{VAE-RNN} & \mbox{Variational Autoencoder built upon RNN} \\
	\mbox{VI} & \mbox{Variational Inference} \\
	\mbox{VRNNaug} & \mbox{Variational RNN via Augmented Recurrent Input Space}
\end{array}
$

\section{Notations} \label{app_notation}
$
\begin{array}{ll}
	\mathcal{B} & \mbox{Subset of short time series chunks} \\
	\mathcal{D} & \mbox{Time series dataset} \\
	\mathcal{D}_c & \mbox{Converted short time series chunks} \\
	\mathcal{D}_c^j & \mbox{$j$-th short time series chunk} \\
	d_{\mbf{u}} & \mbox{Dimensionality of input signal} \\
	{d_{\mbf{z}}} & \mbox{Dimensionality of latent state} \\
	d_{\mbf{h}} & \mbox{Dimensionality of hidden state} \\
	d_{\mbf{y}} & \mbox{Dimensionality of output} \\
	F & \mbox{Forecast horizon} \\
	f & \mbox{Transition function} \\
	g & \mbox{Measurement function} \\
	\mathbf{h}_t & \mbox{Hidden state of RNN or SSM at time $t$} \\
	J & \mbox{Number of short time series chunks} \\
	K & \mbox{Number of Monte Carlo samples} \\
	\mathcal{L} & \mbox{Evidence lower bound calculated on $\mathcal{D}$} \\
	\tilde{\mathcal{L}} & \mbox{Unbiased estimation of evidence lower bound on $\mathcal{D}_c$} \\
	\mathcal{L}^j & \mbox{Evidence lower bound calculated on $\mathcal{D}_c^j$} \\
	\mathcal{M} & \mbox{Sequence model} \\
	\mathcal{N}(.,.) & \mbox{Gaussian distribution} \\
	p50 & \mbox{50\%-quantile loss of future forecasting} \\
	p90 & \mbox{90\%-quantile loss of future forecasting} \\
	\mathrm{QL}_{\rho} & \mbox{$\rho$-quantile loss} \\
	T & \mbox{Sequence length of training time series} \\
	\mathbf{u}_t & \mbox{Input signal at time $t$} \\
	\bar{\mathbf{z}}_t & \mbox{Recurrent latent state until time $t$} \\
	W & \mbox{Chunk size} \\
	\mathbf{y}_t & \mbox{Output observation at time $t$} \\
	\bar{\mathbf{y}}_t & \mbox{Recurrent output until time $t$} \\
	\hat{\mathbf{y}}_t & \mbox{Output sample at time $t$} \\
	\dot{\mathbf{y}}_t & \mbox{Hybrid output at time $t$} \\
	\mathbf{z}_t & \mbox{Latent state at time $t$} \\
	\hat{\mathbf{z}}_t & \mbox{Latent sample at time $t$} \\
	\bar{\mathbf{z}}_t & \mbox{Recurrent latent state until time $t$} \\
	\bm{\epsilon} & \mbox{Random variable following unit Gaussian distribution} \\
	\phi^u(.), \phi^z(.), \phi^y(.) & \mbox{Recurrent mappings} \\
	\bm{\mu}_{\mbf{z}_t}, \bm{\nu}_{\mbf{z}_t}, \bm{\Sigma}_{\mbf{z}_t} & \mbox{Mean, variance and diagonalized variance of $q(\mbf{z}_t|.)$} \\
	\bm{\mu}_{\mbf{y}_t}, \bm{\nu}_{\mbf{y}_t}, \bm{\Sigma}_{\mbf{y}_t} & \mbox{Mean, variance and diagonalized variance of $p(\mbf{y}_t|.)$} \\
	\odot & \mbox{Element-wise product operator}
\end{array}
$

\section{Experimental details} \label{app_exp_details}
\textbf{Toy cases.} For the two toy cases, we adopt the same settings for the following system identification benchmarks except that (i) the chunk size is set as $W=133$ and the latent dimensionality takes $d_{\mbf{z}}=20$ on the \texttt{motorcycle} dataset, and (ii) the Adam optimizer is ran over 200 epochs.

\textbf{System identification benchmarks.} The experimental configurations on the eight system identification datasets ($\mathtt{actuator}$, $\mathtt{ctank}$, $\mathtt{drive}$, $\mathtt{dryer}$, $\mathtt{f16gvt}$, $\mathtt{furnace}$, $\mathtt{tank}$, and $\mathtt{hammerstein}$) are detailed as below. 

Firstly, we perform standardization on the sequence inputs and outputs of time series data to have zero mean and unit variance. We then split the whole data by choosing the first 50\% as training set, the middle 20\% as validation set, and the final 30\% as testing set. Besides, we use the shingling strategy~\cite{leskovec2014mining} to further split the training and validation datasets with the chunk size (i.e., sequence length) $W = 64$, thus resulting into many short time series sets.

As for model configuration of the proposed VRNNaug, we build three GRU models in~\eqref{eq_ar_z}-\eqref{eq_ar_y} using a single hidden layer with 100 units. The weight parameters of these GRU models are initialized by the orthogonal initialization~\cite{saxe2013exact}, and the bias parameters are initialized with zeros. As for the MLPs in VRNNaug, we adopt the fully-connected (FC) neural network with three hidden layers, the ReLU activation, and skip connection~\cite{he2016deep} if possible. The number of units for the hidden layers is set as $\mathrm{max}[ d_{\mbf{x}}, 50]$ wherein $d_{\mbf{x}}$ is the number of input features. Finally, for the VAE structure in Fig.~\ref{fig_vrnnaug}, we set the latent dimensionality of $\mbf{z}$ as $d_{\mbf{z}}=10$.

As for model training, the Adam optimizer is employed with the batch size of $|\mathcal{B}|=128$ and the maximum number of epochs as 100. Particularly, we have a scheduled learning strategy. That is, we start using a learning rate of $1 \times 10^{-3}$ within the first ten epochs, and then check it every ten epochs: if the validation loss cannot decrease within the last ten epochs, we use a decreasing factor 0.5 to adjust the learning rate. We stop the training when (i) it reaches the maximum number of epochs; or (ii) the adjusted learning rate is below $1 \times 10^{-6}$.

As for model forecasting, we start from cold and recursively sample $K=100$ points from the predictive distributions of deep stochastic SSM models at the future time period $[T+1, T+F]$, and then use the prediction samples to evaluate the related $p50$ and $p90$ quantile criteria.

It is finally notable that we repeat the experiments on each time series task ten times by starting with different random seeds in order to comprehensively evaluate the performance of deep sequence models.

\textbf{Centrifugal compressor sensor data.} 
After obtaining the processed \texttt{compressor} sequence data with $d_{\mbf{u}} = 10$ and $d_{\mbf{y}} = 1$, we adopt the similar experimental settings on the above system identification benchmarks for model training and forecasting. The differences are that we here split the whole data by choosing the first 80\% as training set, the middle 10\% as validation set, and the final 10\% as testing set; and we use the shingling strategy to split the training and validation datasets with the chunk size $W = 128$.

\section{Illustration of VRNNaug on benchmarks} \label{app_benchmarks}
\begin{figure*}[t!]
	\centering
	\includegraphics[width=1.0\textwidth]{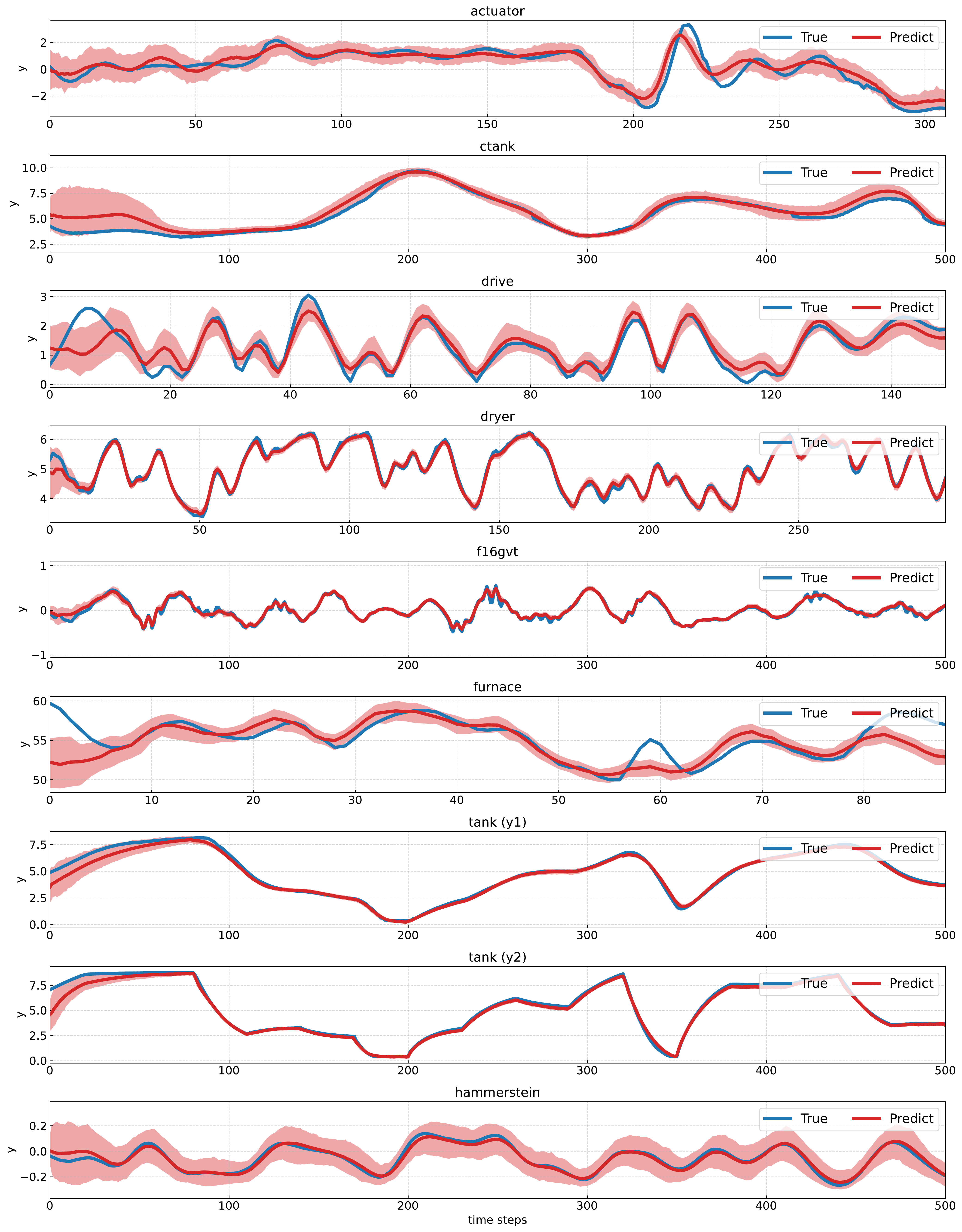}
	\caption{Illustration of the VRNNaug forecasting on eight system identification benchmarks. The red curve represents 50-quantile prediction, while the associated shaded region indicates the interval bounded by 95-quantile and 5-quantile predictions.}
	\label{fig_sysid_vrnnaug} 
\end{figure*}

Fig.~\ref{fig_sysid_vrnnaug} depicts the predictive distributions of the proposed VRNNaug on eight system identification benchmarks, respectively. Note that for the benchmarks with many test time points (e.g., $F \ge 500$), we only illustrate the predictions at the first 500 time points. It is observed that the predictions of VRNNaug well describe the sequence behaviors on most benchmarks.

\end{appendices}

\section*{References}
\bibliography{VRNNaug_ref}

\begin{thebibliography}{10}
\expandafter\ifx\csname url\endcsname\relax
  \def\url#1{\texttt{#1}}\fi
\expandafter\ifx\csname urlprefix\endcsname\relax\def\urlprefix{URL }\fi
\expandafter\ifx\csname href\endcsname\relax
  \def\href#1#2{#2} \def\path#1{#1}\fi

\bibitem{aastrom1971system}
K.~J. {\AA}str{\"o}m, P.~Eykhoff, System identification—{A} survey,
  Automatica 7~(2) (1971) 123--162.

\bibitem{verhaegen2007filtering}
M.~Verhaegen, V.~Verdult, Filtering and system identification: {A} least
  squares approach, Cambridge university press, 2007.

\bibitem{zhao2019deep}
R.~Zhao, R.~Yan, Z.~Chen, K.~Mao, P.~Wang, R.~X. Gao, Deep learning and its
  applications to machine health monitoring, Mechanical Systems and Signal
  Processing 115 (2019) 213--237.

\bibitem{yu2019remaining}
W.~Yu, I.~Y. Kim, C.~Mechefske, Remaining useful life estimation using a
  bidirectional recurrent neural network based autoencoder scheme, Mechanical
  Systems and Signal Processing 129 (2019) 764--780.

\bibitem{lei2020applications}
Y.~Lei, B.~Yang, X.~Jiang, F.~Jia, N.~Li, A.~K. Nandi, Applications of machine
  learning to machine fault diagnosis: {A} review and roadmap, Mechanical
  Systems and Signal Processing 138 (2020) 106587.

\bibitem{tadjer2021machine}
A.~Tadjer, A.~Hong, R.~B. Bratvold, Machine learning based decline curve
  analysis for short-term oil production forecast, Energy Exploration \&
  Exploitation (2021) 01445987211011784.

\bibitem{hyndman2008forecasting}
R.~Hyndman, A.~B. Koehler, J.~K. Ord, R.~D. Snyder, Forecasting with
  exponential smoothing: the state space approach, Springer Science \& Business
  Media, 2008.

\bibitem{box1968some}
G.~E. Box, G.~M. Jenkins, Some recent advances in forecasting and control,
  Journal of the Royal Statistical Society: Series C (Applied Statistics)
  17~(2) (1968) 91--109.

\bibitem{durbin2012time}
J.~Durbin, S.~J. Koopman, Time series analysis by state space methods, Oxford
  university press, 2012.

\bibitem{schon2011system}
T.~B. Sch{\"o}n, A.~Wills, B.~Ninness, System identification of nonlinear
  state-space models, Automatica 47~(1) (2011) 39--49.

\bibitem{hanachi2016sequential}
H.~Hanachi, J.~Liu, A.~Banerjee, Y.~Chen, Sequential state estimation of
  nonlinear/non-gaussian systems with stochastic input for turbine degradation
  estimation, Mechanical Systems and Signal Processing 72 (2016) 32--45.

\bibitem{rabiner1986introduction}
L.~Rabiner, B.~Juang, An introduction to hidden {M}arkov models, IEEE ASSP
  Magazine 3~(1) (1986) 4--16.

\bibitem{liao2018uncertainty}
Y.~Liao, L.~Zhang, C.~Liu, Uncertainty prediction of remaining useful life
  using long short-term memory network based on bootstrap method, in: IEEE
  International Conference on Prognostics and Health Management, IEEE, 2018,
  pp. 1--8.

\bibitem{kalman1960new}
R.~E. Kalman, A new approach to linear filtering and prediction problems,
  Journal of Basic Engineering 82~(1) (1960) 35--45.

\bibitem{west2006bayesian}
M.~West, J.~Harrison, Bayesian forecasting and dynamic models, Springer Science
  \& Business Media, 2006.

\bibitem{williams2006gaussian}
C.~K. Williams, C.~E. Rasmussen, Gaussian processes for machine learning, MIT
  press Cambridge, MA, 2006.

\bibitem{frigola2014variational}
R.~{Frigola}, Y.~{Chen}, C.~{Rasmussen}, Variational {G}aussian process
  state-space models, in: Advances in Neural Information Processing Systems,
  Vol.~27, 2014, pp. 3680--3688.

\bibitem{ialongo2019overcoming}
A.~D. {Ialongo}, M.~van~der {Wilk}, J.~{Hensman}, C.~E. {Rasmussen}, Overcoming
  mean-field approximations in recurrent {G}aussian process models., in:
  International Conference on Machine Learning, 2019, pp. 2931--2940.

\bibitem{liu2020gaussian}
H.~Liu, Y.-S. Ong, X.~Shen, J.~Cai, When gaussian process meets big data: {A}
  review of scalable {GP}s, IEEE Transactions on Neural Networks and Learning
  Systems 31~(11) (2020) 4405--4423.

\bibitem{snelson2006sparse}
E.~Snelson, Z.~Ghahramani, Sparse gaussian processes using pseudo-inputs, in:
  Advances in Neural Information Processing Systems, MIT Press, 2006, pp.
  1257--1264.

\bibitem{titsias2009variational}
M.~Titsias, Variational learning of inducing variables in sparse {G}aussian
  processes, in: Artificial Intelligence and Statistics, 2009, pp. 567--574.

\bibitem{hensman2013gaussian}
J.~Hensman, N.~Fusi, N.~D. Lawrence, Gaussian processes for big data, in:
  Uncertainty in Artificial Intelligence, Citeseer, 2013, p. 282.

\bibitem{liu2018generalized}
H.~Liu, J.~Cai, Y.~Wang, Y.-S. Ong, Generalized robust {B}ayesian committee
  machine for large-scale {G}aussian process regression, in: International
  Conference on Machine Learning, 2018, pp. 1--10.

\bibitem{cho2014learning}
K.~Cho, B.~van Merri{\"e}nboer, C.~Gulcehre, D.~Bahdanau, F.~Bougares,
  H.~Schwenk, Y.~Bengio, Learning phrase representations using {RNN}
  encoder--decoder for statistical machine translation, in: Proceedings of the
  2014 Conference on Empirical Methods in Natural Language Processing, 2014,
  pp. 1724--1734.

\bibitem{hochreiter1997long}
S.~Hochreiter, J.~Schmidhuber, Long short-term memory, Neural Computation 9~(8)
  (1997) 1735--1780.

\bibitem{sutskever2014sequence}
I.~Sutskever, O.~Vinyals, Q.~V. Le, Sequence to sequence learning with neural
  networks, arXiv preprint arXiv:1409.3215.

\bibitem{gehring2017convolutional}
J.~Gehring, M.~Auli, D.~Grangier, D.~Yarats, Y.~N. Dauphin, Convolutional
  sequence to sequence learning, in: International Conference on Machine
  Learning, PMLR, 2017, pp. 1243--1252.

\bibitem{lipton2015critical}
Z.~C. Lipton, J.~Berkowitz, C.~Elkan, A critical review of recurrent neural
  networks for sequence learning, arXiv preprint arXiv:1506.00019.

\bibitem{yu2021analysis}
W.~Yu, I.~Y. Kim, C.~Mechefske, Analysis of different rnn autoencoder variants
  for time series classification and machine prognostics, Mechanical Systems
  and Signal Processing 149 (2021) 107322.

\bibitem{salinas2020deepar}
D.~Salinas, V.~Flunkert, J.~Gasthaus, T.~Januschowski, Deep{AR}:
  {P}robabilistic forecasting with autoregressive recurrent networks,
  International Journal of Forecasting 36~(3) (2020) 1181--1191.

\bibitem{li2021learning}
L.~Li, J.~Yan, X.~Yang, Y.~Jin, Learning interpretable deep state space model
  for probabilistic time series forecasting, arXiv preprint arXiv:2102.00397.

\bibitem{rangapuram2018deep}
S.~S. Rangapuram, M.~W. Seeger, J.~Gasthaus, L.~Stella, Y.~Wang,
  T.~Januschowski, Deep state space models for time series forecasting,
  Advances in Neural Information Processing Systems 31 (2018) 7785--7794.

\bibitem{yanchenko2020stanza}
A.~K. Yanchenko, S.~Mukherjee, Stanza: {A} nonlinear state space model for
  probabilistic inference in non-stationary time series, arXiv preprint
  arXiv:2006.06553.

\bibitem{al2017learning}
M.~Al-Shedivat, A.~G. Wilson, Y.~Saatchi, Z.~Hu, E.~P. Xing, Learning scalable
  deep kernels with recurrent structure, The Journal of Machine Learning
  Research 18~(1) (2017) 2850--2886.

\bibitem{she2020neural}
Q.~She, A.~Wu, Neural dynamics discovery via {G}aussian process recurrent
  neural networks, in: Uncertainty in Artificial Intelligence, PMLR, 2020, pp.
  454--464.

\bibitem{kingma2013auto}
D.~P. Kingma, M.~Welling, Auto-encoding variational {B}ayes, arXiv preprint
  arXiv:1312.6114.

\bibitem{bayer2014learning}
J.~Bayer, C.~Osendorfer, Learning stochastic recurrent networks, arXiv preprint
  arXiv:1411.7610.

\bibitem{krishnan2015deep}
R.~G. Krishnan, U.~Shalit, D.~Sontag, Deep kalman filters, arXiv preprint
  arXiv:1511.05121.

\bibitem{krishnan2017structured}
R.~G. Krishnan, U.~Shalit, D.~Sontag, Structured inference networks for
  nonlinear state space models, in: Proceedings of the Thirty-First AAAI
  Conference on Artificial Intelligence, 2017, pp. 2101--2109.

\bibitem{fraccaro2018deep}
M.~Fraccaro, Deep latent variable models for sequential data, Ph.D. thesis,
  Department of Applied Mathematics and Computer Science, Technical University
  of Denmark (2018).

\bibitem{gedon2020deep}
D.~{Gedon}, N.~{Wahlström}, T.~B. {Schön}, L.~{Ljung}, Deep state space
  models for nonlinear system identification., arxiv:eess.SY.

\bibitem{leskovec2014mining}
J.~Leskovec, A.~Rajaraman, J.~D. Ullman, Mining of massive data sets, Cambridge
  university press, 2014.

\bibitem{kingma2014adam}
D.~P. Kingma, J.~Ba, Adam: {A} method for stochastic optimization, arXiv
  preprint arXiv:1412.6980.

\bibitem{bengio2015scheduled}
S.~Bengio, O.~Vinyals, N.~Jaitly, N.~Shazeer, Scheduled sampling for sequence
  prediction with recurrent neural networks, in: Advances in Neural Information
  Processing Systems, 2015, pp. 1171--1179.

\bibitem{zhang2019bridging}
W.~Zhang, Y.~Feng, F.~Meng, D.~You, Q.~Liu, Bridging the gap between training
  and inference for neural machine translation, in: Proceedings of the 57th
  Annual Meeting of the Association for Computational Linguistics, 2019, pp.
  4334--4343.

\bibitem{bai2018empirical}
S.~Bai, J.~Z. Kolter, V.~Koltun, An empirical evaluation of generic
  convolutional and recurrent networks for sequence modeling, arXiv preprint
  arXiv:1803.01271.

\bibitem{vaswani2017attention}
A.~Vaswani, N.~Shazeer, N.~Parmar, J.~Uszkoreit, L.~Jones, A.~N. Gomez,
  {\L}.~Kaiser, I.~Polosukhin, Attention is all you need, in: Advances in
  Neural Information Processing Systems, 2017, pp. 6000--6010.

\bibitem{huang2017densely}
G.~Huang, Z.~Liu, L.~Van Der~Maaten, K.~Q. Weinberger, Densely connected
  convolutional networks, in: Proceedings of the IEEE Conference on Computer
  Vision and Pattern Recognition, 2017, pp. 4700--4708.

\bibitem{chen2020probabilistic}
Y.~Chen, Y.~Kang, Y.~Chen, Z.~Wang, Probabilistic forecasting with temporal
  convolutional neural network, Neurocomputing 399 (2020) 491--501.

\bibitem{lennart1999system}
L.~Lennart, System identification: {T}heory for the user, PTR Prentice Hall,
  Upper Saddle River, NJ 28.

\bibitem{silverman1985some}
B.~W. Silverman, Some aspects of the spline smoothing approach to
  non-parametric regression curve fitting, Journal of the Royal Statistical
  Society: Series B (Methodological) 47~(1) (1985) 1--21.

\bibitem{schoukens2016cascaded}
M.~Schoukens, P.~Mattson, T.~Wigren, J.-P. Noel, Cascaded tanks benchmark
  combining soft and hard nonlinearities, in: Workshop on nonlinear system
  identification benchmarks, 2016, pp. 20--23.

\bibitem{noel2017f}
J.-P. No{\"e}l, M.~Schoukens, F-16 aircraft benchmark based on ground vibration
  test data, in: 2017 Workshop on Nonlinear System Identification Benchmarks,
  2017, pp. 19--23.

\bibitem{schoukens2016wiener}
M.~Schoukens, J.-P. Noel, Wiener-hammerstein benchmark with process noise, in:
  Workshop on nonlinear system identification benchmarks, 2016, pp. 15--19.

\bibitem{xu2016bayesian}
X.~Shengli, J.~Xiaomo, H.~Jinzhi, Y.~Shuhua, W.~Xiaofang, Bayesian wavelet pca
  methodology for turbomachinery damage diagnosis under uncertainty, Mechanical
  Systems and Signal Processing 80~(Dec) (2016) 1--18.

\bibitem{daubechies1992ten}
I.~Daubechies, Ten lectures on wavelets, SIAM, 1992.

\bibitem{saxe2013exact}
A.~M. Saxe, J.~L. McClelland, S.~Ganguli, Exact solutions to the nonlinear
  dynamics of learning in deep linear neural networks, arXiv preprint
  arXiv:1312.6120.

\bibitem{he2016deep}
K.~He, X.~Zhang, S.~Ren, J.~Sun, Deep residual learning for image recognition,
  in: Proceedings of the IEEE conference on Computer Vision and Pattern
  Recognition, 2016, pp. 770--778.

\end{thebibliography}

\end{document}